\pdfoutput=1
\documentclass{article}

\usepackage[preprint]{neurips_2024}

\usepackage[utf8]{inputenc} 
\usepackage[T1]{fontenc}    
\usepackage{hyperref}       
\usepackage{url}            
\usepackage{booktabs}       
\usepackage{amsfonts}       
\usepackage{nicefrac}       
\usepackage{microtype}      
\usepackage[dvipsnames,table,usenames]{xcolor}

\usepackage{natbib}
\usepackage{amsfonts}
\usepackage{amssymb}
\usepackage[usenames,dvipsnames]{xcolor}
\usepackage{colortbl}
\usepackage{caption}
\usepackage{newfloat}

\usepackage{multirow}
\usepackage{graphicx}
\usepackage{subscript}
\usepackage{subcaption}
\usepackage{marvosym}
\usepackage{float}

\definecolor{mygray}{RGB}{243, 248, 255}

\def\hf{\raisebox{-2.5pt}{\includegraphics[height=1.55em]{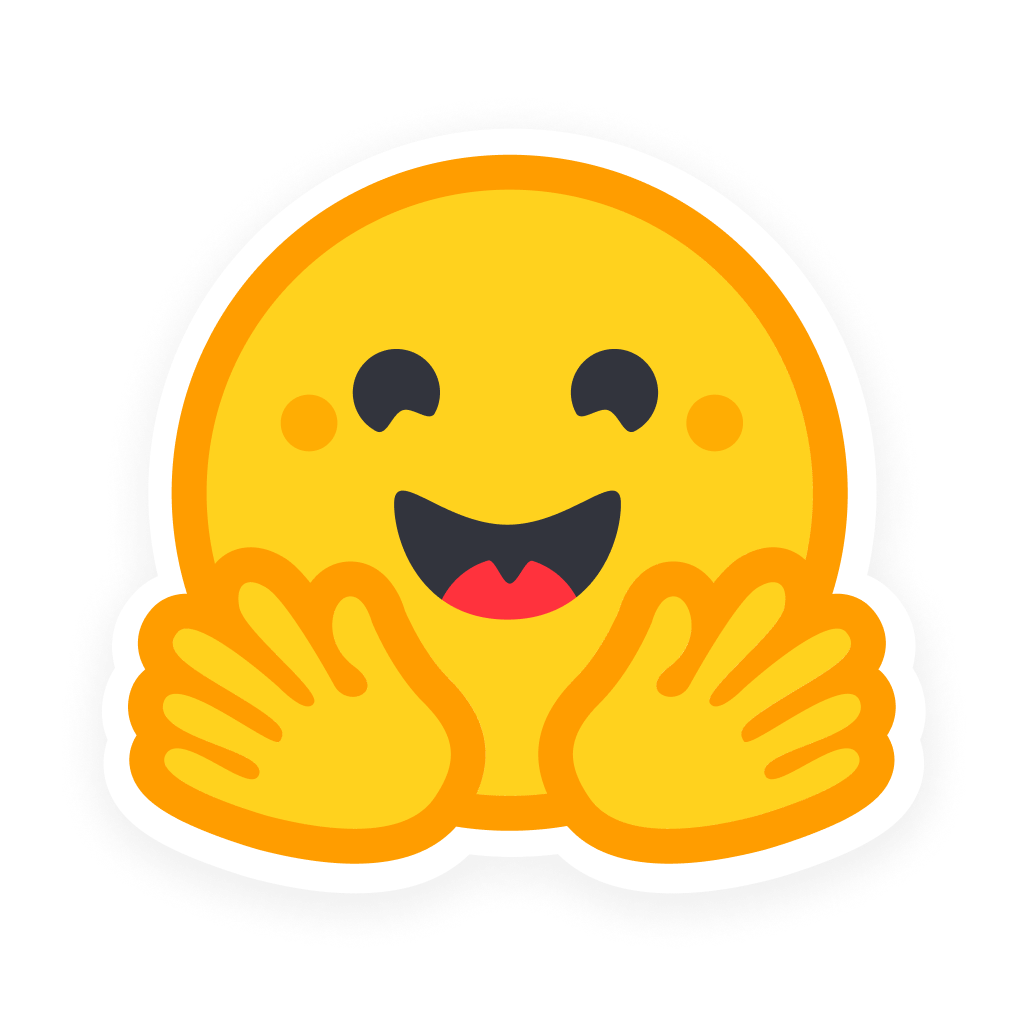}}}
\def\github{\raisebox{-2.5pt}{\includegraphics[height=1.4em]{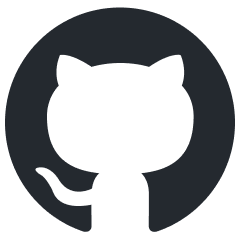}}}
\newcommand{\hflink}{\href{https://huggingface.co/HIT-TMG/KaLM-embedding-multilingual-mini-instruct-v1.5}{https://huggingface.co/HIT-TMG/KaLM-embedding-multilingual-mini-instruct-v1.5}}
\newcommand{\githublink}{\href{https://github.com/HITsz-TMG/KaLM-Embedding}{https://github.com/HITsz-TMG/KaLM-Embedding}}

\title{\textsf{KaLM-Embedding}: Superior Training Data Brings A Stronger Embedding Model}

\author{%
    \textbf{Xinshuo Hu$^{1}$, Zifei Shan, Xinping Zhao$^{1}$, Zetian Sun$^{1}$, Zhenyu Liu$^{1}$, Dongfang Li$^{1}$, } \\
    \textbf{Shaolin Ye, Xinyuan Wei, Qian Chen, Baotian Hu$^{1}$ \textsuperscript{\Letter}, Haofen Wang$^{2}$, Jun Yu$^{1}$, Min Zhang$^{1}$} \\
    $^{1}$Harbin Institute of Technology (Shenzhen), Shenzhen, China \\
    $^{2}$Tongji University, Shanghai, China \\
    \texttt{yanshek.woo@gmail.com, zifeishan@gmail.com, hubaotian@hit.edu.cn} \\
    \hspace{-0.3cm} \hf \quad \hflink \\
    \hspace{-0.3cm} \github \quad \githublink \\
}

\begin{document}

\maketitle

\begin{abstract}
As retrieval-augmented generation prevails in large language models, embedding models are becoming increasingly crucial. 
Despite the growing number of general embedding models, prior work often overlooks the critical role of training data quality. In this work, we introduce \textsf{KaLM-Embedding}, a general multilingual embedding model that leverages a large quantity of cleaner, more diverse, and domain-specific training data.
Our model has been trained with key techniques proven to enhance performance: 
(1) persona-based synthetic data to create diversified examples distilled from LLMs,
(2) ranking consistency filtering to remove less informative samples, 
and (3) semi-homogeneous task batch sampling to improve training efficacy.
Departing from traditional BERT-like architectures, we adopt Qwen2-0.5B as the pre-trained model, facilitating the adaptation of auto-regressive language models for general embedding tasks.
Extensive evaluations of the MTEB benchmark across multiple languages show that our model outperforms others of comparable size, setting a new standard for multilingual embedding models with less than 1B parameters.
\end{abstract}

\section{Introduction}

In recent years, retrieval-augmented generation (RAG) has gained increasing popularity~\citep{DBLP:journals/corr/abs-2312-10997,DBLP:journals/corr/abs-2404-10981}. With the rapid advancement of large language models (LLMs), retrieval models have become the primary bottleneck for improvement within the RAG framework~\citep{DBLP:journals/corr/abs-2404-07221}, leading to the emergence of numerous text embedding models~\citep{DBLP:journals/corr/abs-2405-06932,DBLP:journals/corr/abs-2405-17428,DBLP:journals/corr/abs-2308-03281,DBLP:conf/sigir/XiaoLZMLN24}. As the foundation for information acquisition in RAG systems, a general embedding model is required to demonstrate capabilities across multiple languages, domains, and tasks~\citep{DBLP:conf/eacl/MuennighoffTMR23,DBLP:conf/sigir/XiaoLZMLN24}.

Although numerous general embedding models have been developed using extensive paired data, they frequently overlook the quality of the training data. Specifically, (1) the presence of false negative samples in the fine-tuning data, which are sometimes similar to the positive documents, can introduce noise into representation learning; and (2) the recent success of scaling LLMs demonstrates the promise of cleaner and diverse training data, these aspects remain underemphasized in the development of embedding models.
Therefore, our objective is to develop a superior embedding model by optimizing the quality of the training data and distilling the \textbf{K}nowledge in l\textbf{a}rge \textbf{L}anguage \textbf{M}odels into \textbf{Embedding} Models (\textsf{KaLM-Embedding}).

In this work, we introduce the technical details of our general multilingual text embedding model, \textsf{KaLM-Embedding}. It is adapted from the auto-regressive language model Qwen2-0.5B~\citep{DBLP:journals/corr/abs-2407-10671}, with a carefully crafted training dataset for textual embeddings.
We collect over 20 categories of data for pre-training and 70 categories of data for fine-tuning.
The dataset is further augmented and cleaned with key techniques that we find particularly powerful: 
(1) Persona-based Synthetic Data~\citep{DBLP:conf/acl/WangYHYMW24} to distill LLM knowledge into diverse data for embedding training, 
(2) Ranking Consistency Filtering~\citep{DBLP:conf/iclr/DaiZMLNLBGHC23,DBLP:journals/corr/abs-2212-03533} to improve data quality and reduce noise,
and (3) semi-homogeneous task batching~\citep{SFRAIResearch2024,DBLP:journals/corr/abs-2402-09906}, that combines the same task and random samples for in-batch negatives, to balance the hardness of negatives and the risk of false negatives. 
By combining these useful practices and scaling them into a larger and superior dataset, we demonstrate an even stronger capability for a compact decoder-only embedding model.
Massive evaluation~\citep{DBLP:conf/eacl/MuennighoffTMR23,DBLP:conf/sigir/XiaoLZMLN24,ciancone2024mteb,DBLP:journals/corr/abs-2405-10138} represents that our models achieve state-of-the-art multilingual performance under 0.5B model size.

\section{Training Methodology}

\subsection{Data Collection}

Given our utilization of a pre-trained language model with robust performance capabilities, we adopt the conventional two-stage training approach to develop the embeddings. This process involves weakly-supervised contrastive pre-training followed by supervised fine-tuning.


\begin{figure}[t]
    \centering
    \includegraphics[width=0.8\textwidth]{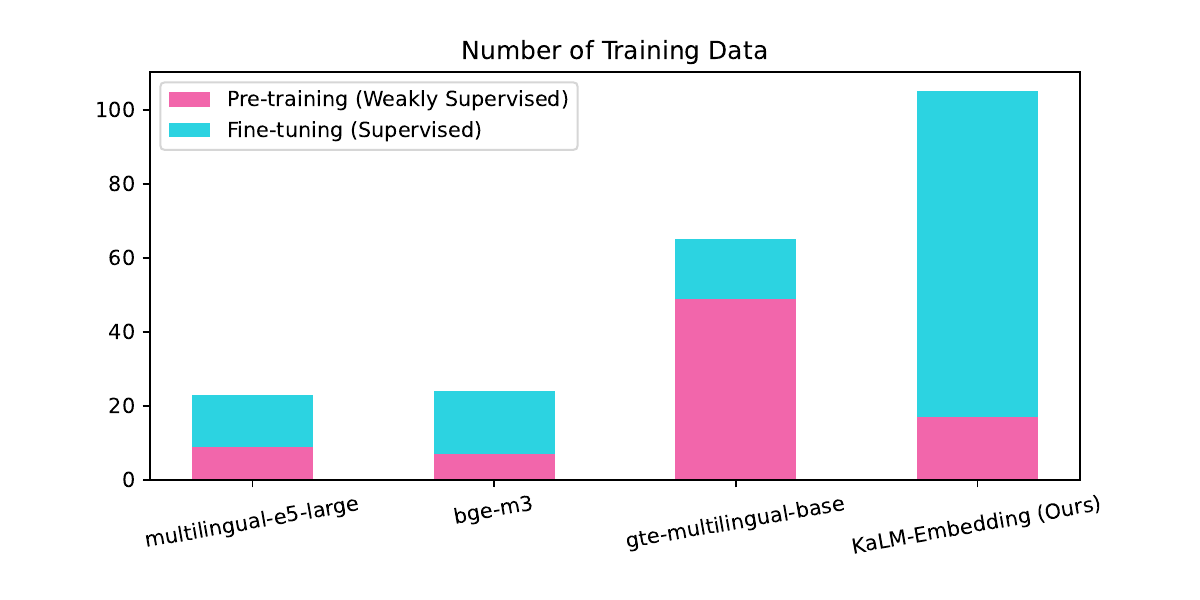}
    \caption{The size of our pre-training and fine-tuning dataset compared to prior work.}
    \label{fig:training_data_comparison_bar}
\end{figure}

\paragraph{Massive Open-source Dataset}  
For the pre-training data, we utilized an extensive collection of title-body pairs from various documents, supplemented by a subset of large-scale supervised question-answering datasets. A detailed list of these datasets is presented in \autoref{tab:pretrain_data_list}.

To ensure the model's generalization capability beyond merely fitting evaluation benchmarks, we employed fine-tuning data from a diverse array of sources. Specifically, during the fine-tuning phase, we incorporated over 70 different datasets, marking a significant departure from previous work in terms of training data selection strategy. These supervised datasets, characterized by high quality and diversity, typically feature relatively small data volumes, making them ideal for fine-tuning in the second phase. A comparison of the number of training datasets is shown in \autoref{fig:training_data_comparison_bar}, and a comprehensive list of the datasets used is provided in \autoref{tab:Fine-tuning_data_list}.

We included several classification and clustering datasets, treating each (sentence, category label) pair as a training instance. For datasets such as Arxiv, we mapped the label abbreviations back to their full label phrases to ensure that the labels convey better semantics. Additionally, we sampled hard negatives from the labels of all classification datasets, rather than mining them through the model. This approach helps mitigate the issue of having too few label categories in certain individual datasets, such as sentiment classification datasets with only two categories. For each specific dataset, we performed minor processing, such as filtering out samples with excessively short documents or excluding lower-quality parts based on the metadata provided with the dataset.

To ensure data integrity and avoid contamination, we only utilized the training sets of all datasets, explicitly excluding any test sets. For some collected classification and clustering datasets without separated training and test sets, we first filtered out the test set samples included in the MTEB from the data, and then processed and sampled the remaining data. This approach guarantees that all examples present in the MTEB evaluations remain unseen during training.

Despite our fine-tuning data being primarily in Chinese and English, with only a small portion of multilingual data, the performance of our model in other languages remains satisfactory, showing that the multi-lingual advantage of pre-trained LLMs can carry over to embedding models.

\paragraph{Persona-based Synthetic Data}  
Following previous work~\citep{DBLP:conf/acl/WangYHYMW24}, we generated 550k synthetic data using large language models, specifically QWen2-72B-Instruct, encompassing 6 types of tasks with 40k unique instructions. To enhance the diversity of the generated data~\citep{DBLP:conf/emnlp/TanLWBJBKL0024}, we incorporated randomly sampled personas from Persona Hub~\citep{DBLP:journals/corr/abs-2406-20094} as the system prompt for the large language model, effectively increasing the domain diversity of the generated data. Since 4 types of retrieval tasks require generating instructions before generating data, we introduced persona roles only during the instruction generation phase to avoid conflicts in persona roles between the two stages.

\begin{figure}[t]
    \centering
    \begin{subfigure}[b]{0.76\textwidth}
        \centering
        \includegraphics[width=\textwidth]{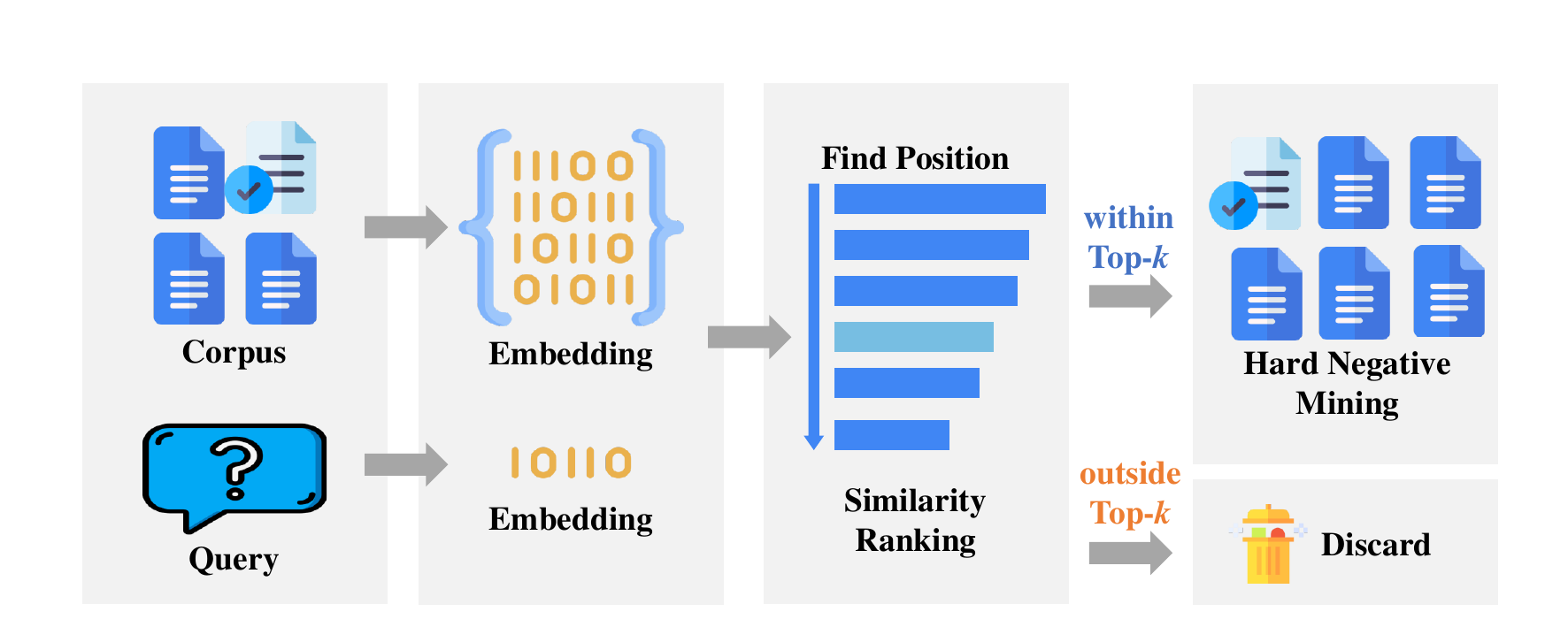}
    \end{subfigure}
    \hfill
    \begin{subfigure}[b]{0.23\textwidth}
        \centering
        \includegraphics[width=\textwidth]{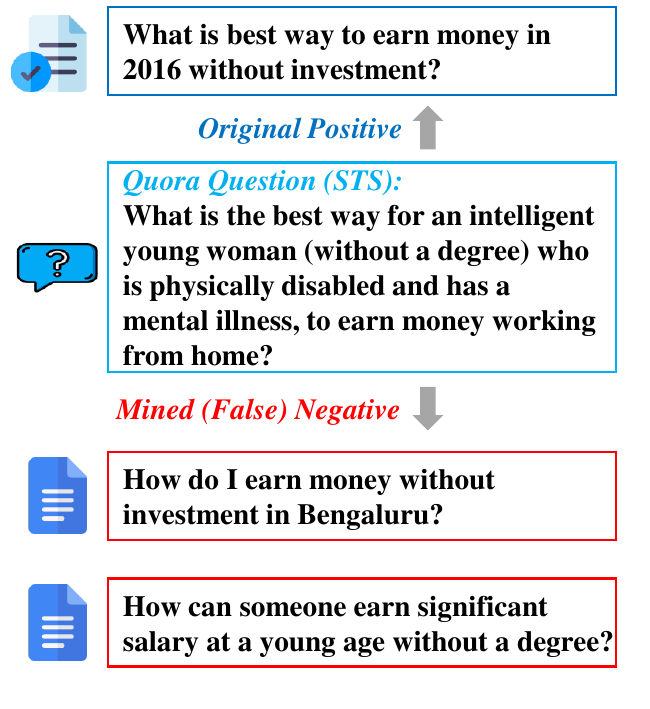}
    \end{subfigure}
    \caption{The framework of ranking consistency filtering. }
    \label{fig:ranking_consistency_filtering}
\end{figure}

\paragraph{Ranking Consistency Filtering}
In addition to utilizing in-batch negatives, another approach for obtaining negative samples involves retrieving hard negatives from the dataset's corpus. However, in some datasets, a single query may correspond to multiple correct documents or answers, 
as exemplified on the right side of \autoref{fig:ranking_consistency_filtering}. 
Furthermore, some queries may be overly broad, leading to their association with multiple documents despite low relevance between the query and the documents. 
Such scenarios can introduce false negatives, which can adversely affect model optimization, whether they are hard negatives (in the former case) or in-batch negatives (in the latter case).

To address this issue, we employ a ranking consistency filtering method, also known as top-k filtering~\citep{DBLP:conf/iclr/DaiZMLNLBGHC23,DBLP:journals/corr/abs-2212-03533}, for fine-tuning data selection. This method involves ranking the similarity between the query and its original positive example data within the entire document corpus of the dataset, and filtering out samples where the positive example data pair does not rank within the top-k. It is important to note that this filtering method is conducted simultaneously with hard negative mining, as both processes require encoding all queries and the document corpus and calculating their relevance. Performing these tasks concurrently can effectively avoid redundant computations. The general process is illustrated on the left side of \autoref{fig:ranking_consistency_filtering}.

\subsection{Training Strategy}

\paragraph{Semi-homogeneous Task Batching} 
Previous studies have employed a method called homogeneous task batching~\citep{SFRAIResearch2024,DBLP:journals/corr/abs-2402-09906}, where batches are formed exclusively from samples belonging to a single task. This approach effectively enhances training efficiency by increasing the hardness of in-batch negatives. Nevertheless, as mentioned earlier, utilizing a large homogeneous task batch carries the substantial drawback of potentially containing an excessive number of false negatives.

\begin{figure}[t]
    \centering
    \includegraphics[width=0.8\textwidth]{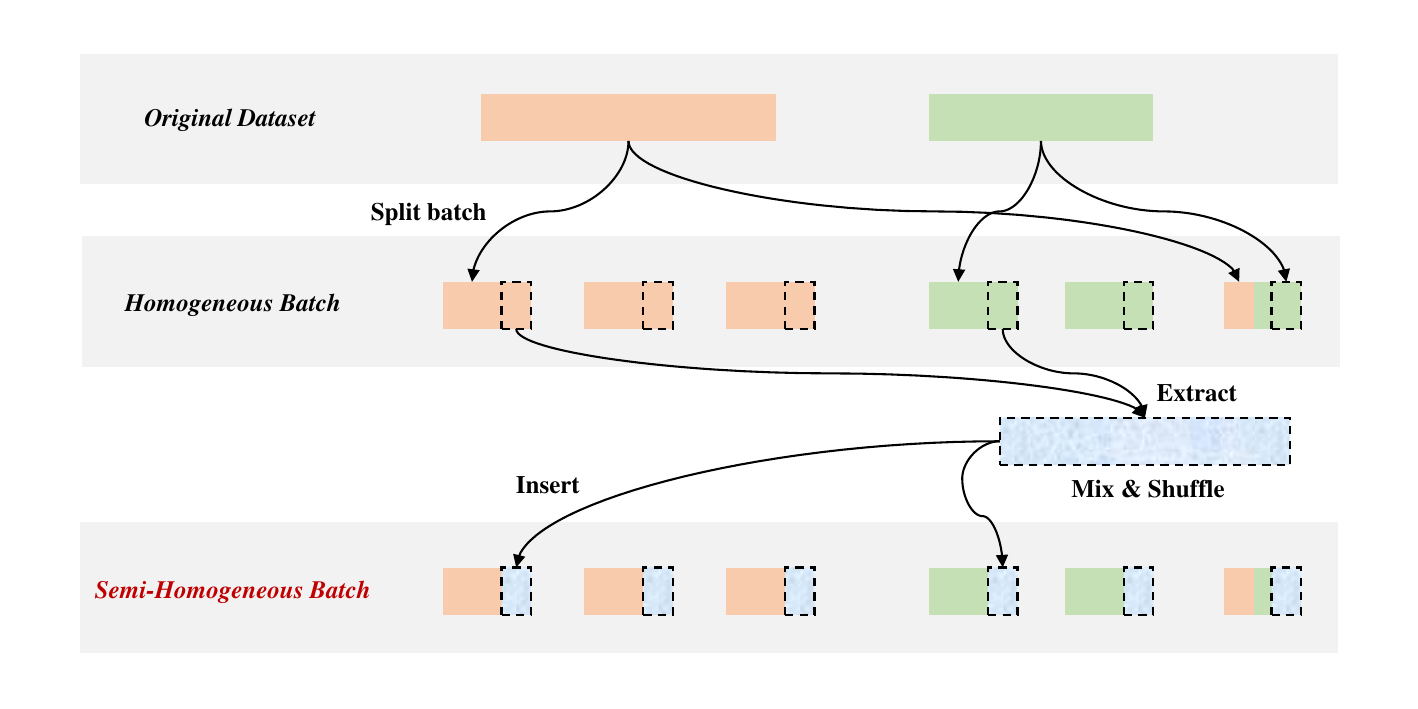}
    \caption{The process to construct the semi-homogeneous task batches.}
    \label{fig:semi-homogeneous-task-batch}
\end{figure}

In this study, we introduce the concept of semi-homogeneous task batching. This method involves creating a semi-batch that includes single-task samples, with the remaining portion of the batch consisting of randomly selected samples. As depicted in \autoref{fig:semi-homogeneous-task-batch}, the semi-homogeneous task batching is constructed as follows: (1) Homogeneous task batch division: initially, a complete homogeneous task batch is constructed; (2) Semi-batch sampling and recombination: subsequently, a specified proportion of these homogeneous batches is sampled, mixed, and randomly reassigned back to their original batches. This approach aims to balance the difficulty of in-batch negatives with the risk of encountering false negatives.

However, this method was NOT utilized in our latest model; nonetheless, it offers a means of controlled analysis. Detailed analysis and conclusions can be found in the subsequent experimental sections.

\paragraph{Matryoshka Representation Learning}
Following many existing models and API services, in order to enhance the utility of embeddings, we have also incorporated Matryoshka Representation Learning (MRL)~\citep{DBLP:conf/nips/KusupatiBRWSRHC22} training to achieve flexible dimension embedding. Specifically, we set the vector dimensions for MRL training to 896, 512, 256, 128, and 64, with corresponding training loss weights of 1.0, 0.3, 0.2, 0.1, and 0.1, respectively.

\paragraph{Task Instruction}
Instruction-finetuned language models are known for their strong generalization capabilities, yet this strategy remains underutilized in embedding models. We find that verbalized instructions can significantly enhance the performance of embedding models by reducing ambiguity in the embedding space.
Concretely, in our training process, we prepended instruction prefixes to the queries of open-source data to differentiate between various tasks, employing a similar setup during testing. The instructions for different tasks are illustrated in \autoref{tab:task_instruction}. For synthetic data, we preserved the originally generated instructions, encompassing a variety of directives for retrieval tasks. Consequently, when deploying the model in practical applications, it is advisable to tailor task instructions to specific scenarios and requirements. Given that our model has been trained on a substantial volume of synthetic instructions, it demonstrates a robust capacity to comprehend and generalize instructions. The instructions used for the Classification and Clustering tasks in the MTEB evaluation are provided in \autoref{tab:task_instruction_detailed_list}.

\begin{table}[htbp]
  \centering
  \footnotesize
  \begin{tabular}{lllp{5.4cm}}
    \toprule
    \multicolumn{2}{c}{\textbf{Task Type}} & \textbf{Instruction} & \textbf{Example} \\
    \hline
    
    \multirow{5}{*}{Asymmetric}  & Retrieval, Reranking & General & Instruct: Given a query, retrieve documents that answer the query. \textbackslash n Query:  \{query\} \\
    \cmidrule(r){2-4}
    
    & Classification, Clustering & Specific & Instruct: Classifying the sentiment expressed in the given movie review text from the IMDB dataset \textbackslash n Query:  \{query\}\\
    
    \hline
    Symmetric & STS, PariClassification & None & - \\
    
    \bottomrule
  \end{tabular}
  \vspace{1mm}
  \caption{The task instruction of query for training and evaluation.}
  \label{tab:task_instruction}
\end{table}

\section{Experiment}

\subsection{Experimental Setting}

We utilize the InfoNCE loss function as our optimization objective, with the temperature parameter set to 0.01. Our base model is Qwen2-0.5B, employing mean pooling. The maximum input text length is capped at 512 tokens. The model is trained using mixed precision with bf16.

For pre-training, we exclusively use in-batch negatives to enhance efficiency. The pre-training process is conducted on 6 nodes, each equipped with 8 Ascend 910B NPUs having 65GB of memory. The model undergoes pre-training for 1 epoch, equivalent to approximately 19k steps, with a batch size of 512 and a learning rate of 1e-4.

During fine-tuning, we incorporate hard negatives after a filtering process. For hard negative mining within the fine-tuning dataset, we sample 7 negative examples from a range of 50 to 100. The top-k threshold for ranking consistency filtering is set to 50. Fine-tuning is performed on 3 nodes with NPUs. The model is fine-tuned for 1 epoch, approximately 4.5k steps, with a batch size of 48 and a learning rate of 1e-4. Matryoshka representation learning is applied during fine-tuning, utilizing dimensions of 896, 512, 256, 128, and 64 as previously specified.

\subsection{Results}

\begin{table}[t]
  \centering
  \footnotesize
  \begin{tabular}{lc|ccccc}
    \toprule
    \multirow{2}{*}{\textbf{Model}} & \multirow{2}{*}{\textbf{Size}} & \multicolumn{5}{c}{\textbf{MTEB}}  \\
    \cmidrule(r){3-7}
               & & \textbf{~ZH~} & \textbf{~EN~} & \textbf{~FR~} & \textbf{~PL~} & \textbf{~avg~} \\
    \hline
    Cohere-embed-multilingual-v3.0  & - & - & 64.01 & 56.02 & - \\
    jina-embeddings-v3 (Multi-LoRA) & 572M & - & 65.51 & 62.29 & 63.97 & - \\
    e5-mistral-7b-instruct          & 7111M & 60.89 & 66.40 & 48.33 & - & -\\ 
    \hline

    paraphrase-multilingual-mpnet-base-v2 & 278M & 44.59 & 54.64 & 55.21 & 48.67 & 50.78 \\
    multilingual-e5-large$^{\dag}$             & 560M & 58.54 & 60.89 & 55.64 & 60.08 & 58.79 \\
    bge-m3 (Dense)$^{\dag}$                    & 560M & 61.07 & 59.57 & 58.79 & \textbf{60.35} & 59.95 \\
    gte-multilingual-base (Dense)$^{\dag}$    & 305M & 62.72 & 61.40 & 59.79 & 58.22 & 60.53 \\
    \rowcolor{mygray}
    \textbf{KaLM-embedding-mini-instruct}   & 494M & \textbf{64.13} & \textbf{64.94} & \textbf{63.08} & 57.05 & \textbf{62.3} \\
    \bottomrule
  \end{tabular}
  \vspace{1mm}
  \caption{Evaluation results on MTEB Chinese, English, French and Polish\protect\footnotemark. Our model, \textbf{KaLM-embedding-mini-instruct}, achieves a new state-of-the-art among multilingual embedding models of <1B parameters, an economical choice for building applications such as retrieval-augmented systems.}
  \label{tab:mteb_main_results}
\end{table}
\footnotetext{The $\dag$ symbol indicates results reused from \citet{DBLP:conf/emnlp/ZhangZLXDTLYXHZ24}. Other evaluation results are sourced from the \href{https://huggingface.co/spaces/mteb/leaderboard}{MTEB leaderboard} (accessed on December 25, 2024).}

\paragraph{MTEB Evaluation:} We employed the Massive Text Embedding Benchmark (MTEB)~\citep{DBLP:conf/eacl/MuennighoffTMR23,DBLP:conf/sigir/XiaoLZMLN24,ciancone2024mteb,DBLP:journals/corr/abs-2405-10138} as the primary dataset for evaluation and analysis due to its diverse task types and extensive datasets. 
While our primary optimization targets were Chinese (zh) and English (en), we also conducted evaluations in French (fr) and Polish (pl). A summary of the multilingual evaluation results is presented in \autoref{tab:mteb_main_results}, with detailed results for each language provided in the appendix. Our KaLM-embedding-mini-instruct model demonstrated significantly superior overall performance across multiple languages compared to other models. However, its performance in Polish was relatively weaker, likely due to the lower proportion of Polish in the training data, particularly within the language distribution of synthetic data~\citep{DBLP:conf/acl/ConneauKGCWGGOZ20,DBLP:conf/acl/WangYHYMW24}.

\begin{table}[t]
  \centering
  \footnotesize
  \begin{tabular}{lcc}
    \toprule
    \multirow{2}{*}{\textbf{Model}} & \multicolumn{2}{c}{\textbf{MTEB}}  \\
    \cmidrule(r){2-3}
               & \textbf{~~~~ZH~~~~} & \textbf{~~~~EN~~~~} \\
    \hline

    \textbf{KaLM-embedding-mini-instruct}  & 64.13 & 64.94 \\
    \quad w/o Matryoshka Representation Learning & 64.07 & 65.00 \\
    \quad w/o Weakly-supervised Pre-training & 64.06 & 64.07 \\
    \quad w/o Task Instructions    & 61.57 & 61.13 \\
    \quad w/o Ranking Consistency Filtering & 64.25 & 64.19 \\

    \bottomrule
  \end{tabular}
  \vspace{1mm}
  \caption{Ablation of different training strategies.}
  \label{tab:ablation_study}
\end{table}

\paragraph{Ablation Study:} We conducted ablation studies on training strategies and data filtering, with the experimental results shown in \autoref{tab:ablation_study}. Since we assigned a smaller weight to the low-dimensional Matryoshka embedding during training, the Matryoshka Representation Learning had a minimal impact on the final results. An analysis of results across different dimensions is provided in subsequent sections. The impact of task instruction was particularly significant, especially given our use of a mixture of various types of training data. Ranking consistency filtering was particularly effective for our previous models, but its impact was relatively minor on this latest model, possibly due to the more comprehensive data coverage in this version. A similar ablation effect was observed with pre-training, which contrasts with the findings of many previous studies~\citep{DBLP:journals/corr/abs-2402-03216,DBLP:conf/emnlp/ZhangZLXDTLYXHZ24}. Additionally, data filtering had a more pronounced improvement effect on English than on Chinese, likely because the English evaluation included more out-of-domain data, making the enhancement in generalization through data filtering more evident in English.

\subsection{Analysis}

\begin{figure}[t]
    \centering
    \begin{subfigure}[b]{0.495\textwidth}
        \centering
        \includegraphics[width=\textwidth]{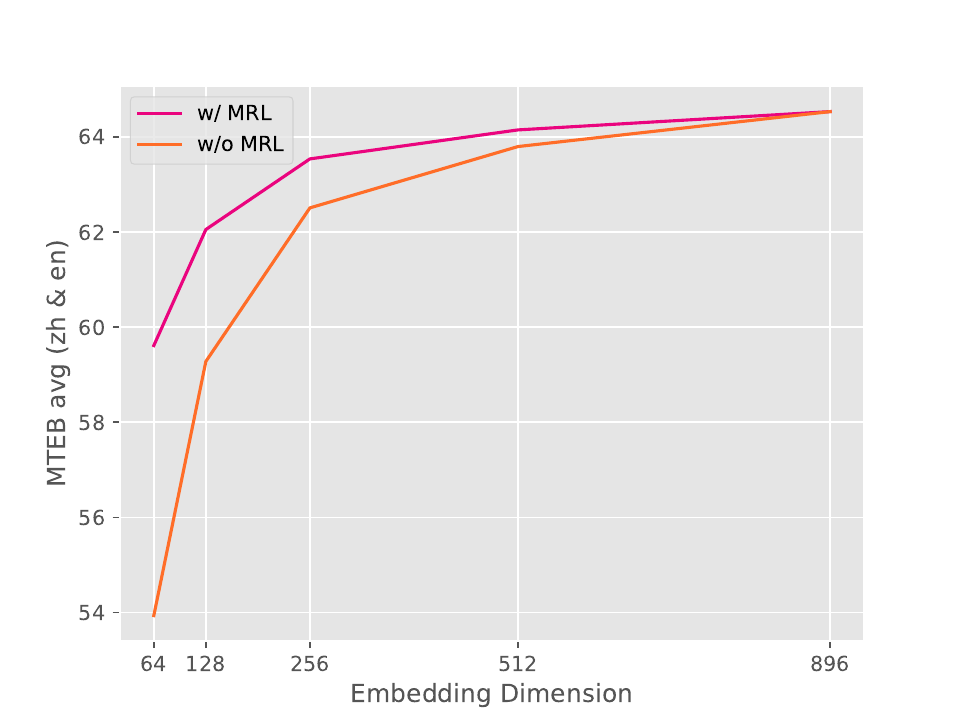}
        \subcaption{Matryoshka Representation Learning.}
        \label{fig:Matryoshka_Representation_Learning}
    \end{subfigure}
    \hfill
    \begin{subfigure}[b]{0.495\textwidth}
        \centering
        \includegraphics[width=\textwidth]{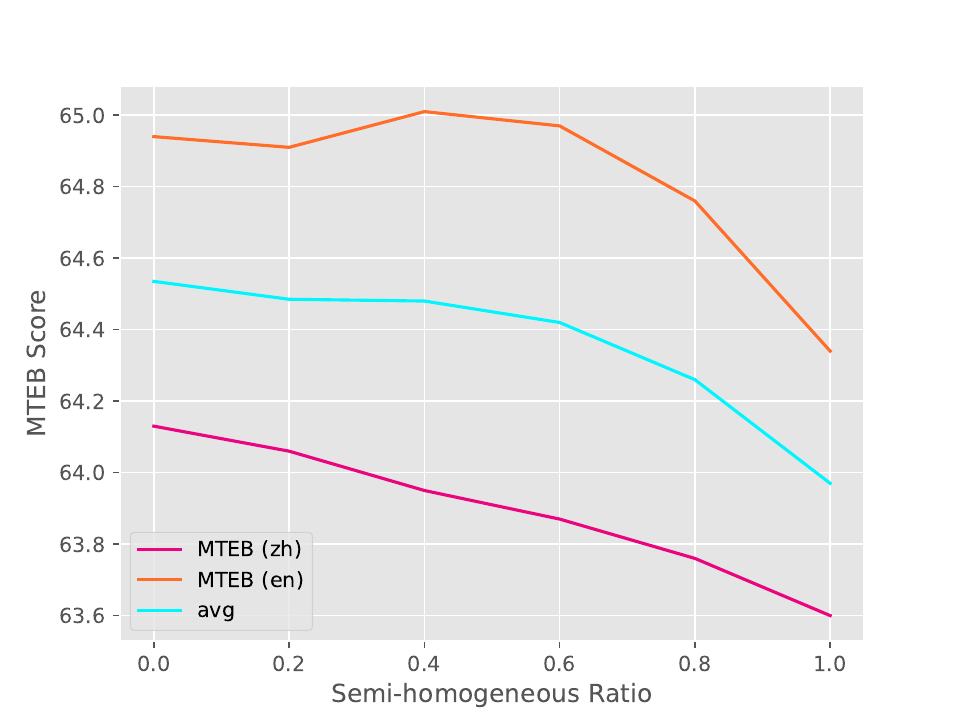}
        \subcaption{Semi-homogeneous Task Batch}
        \label{fig:Semi-homogeneous_Task_Batch}
    \end{subfigure}
    \caption{Impact of training strategy and parameters on MTEB in Chinese and English.}
    \label{fig:analysis_trend}
\end{figure}

\paragraph{Matryoshka Embedding:} We assessed the performance of embedding vectors truncated to various dimensions, as depicted in \autoref{fig:Matryoshka_Representation_Learning}. The results clearly indicate that performance deteriorates with decreasing dimensionality. Nonetheless, models trained using Matryoshka Representation Learning demonstrate substantial improvements in the performance of low-dimensional embeddings. This enhancement is likely limited by our configuration of relatively small learning weights for low-dimensional embeddings, implying that there remains significant potential for further improvement in the performance of low-dimensional embeddings relative to their full-dimensional counterparts.

\paragraph{Semi-homogeneous Task Batch:} In \autoref{fig:Semi-homogeneous_Task_Batch}, we investigate the impact of varying ratios of Semi-homogeneous Task Batches on the final outcomes. Our analysis reveals that increasing the proportion of Semi-homogeneous Tasks negatively affects overall performance. As depicted in \autoref{fig:semi-homogeneous_task_batch_trend_detailed_tasks} in the appendix, this detrimental effect is particularly evident in classification, clustering, and pair-classification tasks, whereas it positively influences retrieval and reranking tasks. This discrepancy primarily arises because the labels in the classification and clustering data within the training set are generally limited, leading to a significant number of false in-batch negatives when homogeneous task batches are used, especially when the batch size is large. Although this training strategy was not adopted in our final model, it remains a viable optimization method depending on specific optimization goals and the nature of the training data. This approach could be particularly useful in scenarios where classification/clustering training data is not utilized, or when there is a specific aim to optimize retrieval and reranking tasks.

\begin{figure}[t]
    \centering
    \includegraphics[width=0.8\textwidth]{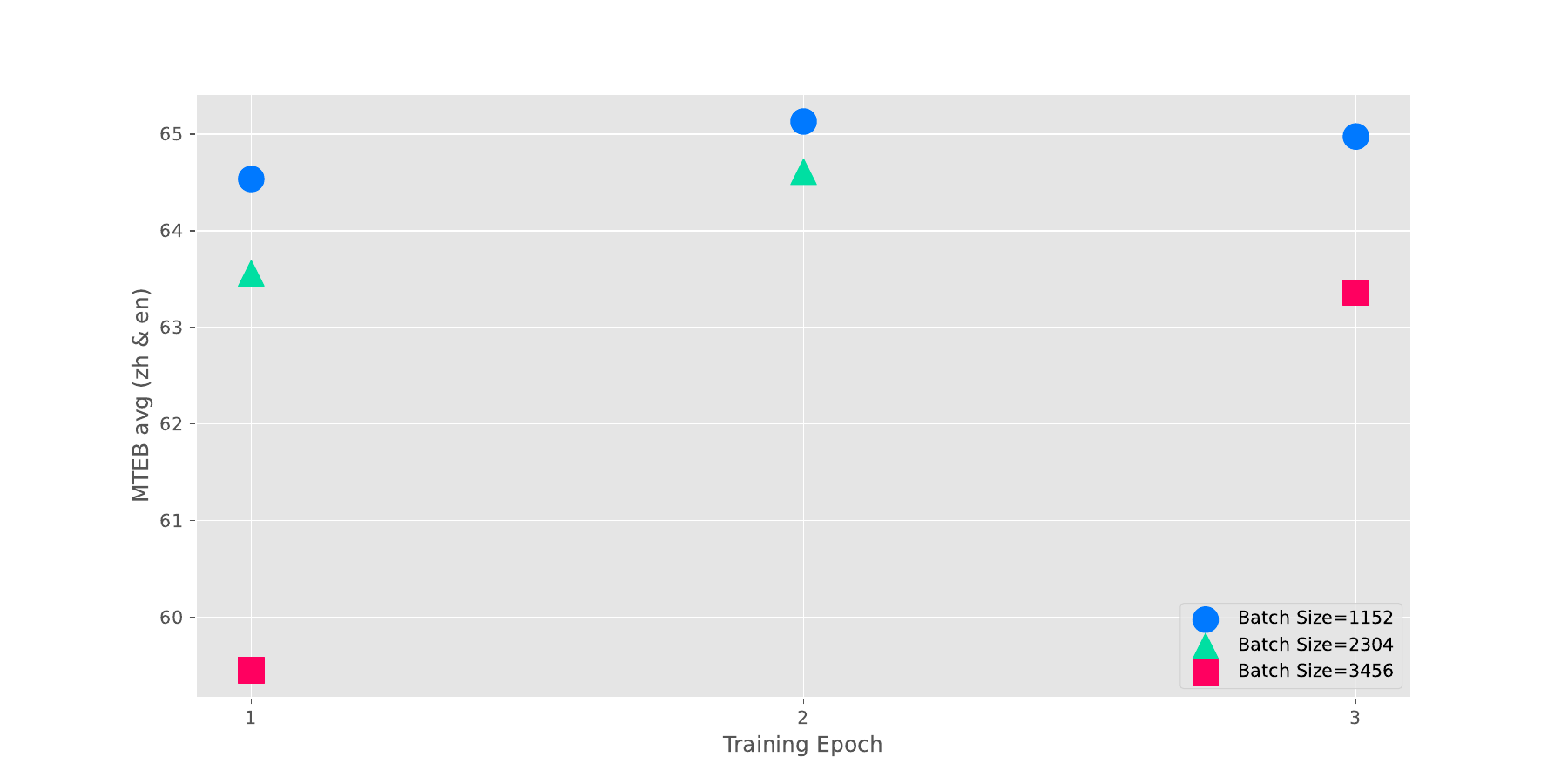}
    \caption{The impact of training epochs and batch size on MTEB in Chinese and English.}
    \label{fig:batch_step_scatter}
\end{figure}

\paragraph{Step and Batch Size:} We conducted an analysis to examine the impact of training epochs (or training steps) and batch size (or the number of training nodes) on model performance, as illustrated in \autoref{fig:batch_step_scatter}. The batch size was regulated by adjusting the number of training nodes, with a focus on the global batch size, excluding hard negatives. The experiments were performed using 3, 6, and 9 nodes. Given that training for one epoch with varying batch sizes results in different numbers of training steps for the same volume of training data, we also compared the outcomes across different epochs. 
Our findings indicate that an increased number of training steps positively influenced the overall performance of the model on the MTEB benchmark. However, when comparing scenarios with an equivalent number of steps (1 epoch with 3 nodes, 2 epochs with 6 nodes, and 3 epochs with 9 nodes), a larger batch size did not necessarily lead to better results. This phenomenon could be attributed to the higher probability of false negatives within the batch as the batch size increases, which also resulted in greater fluctuations in the observed loss during training. 
Ultimately, we selected the model trained with 1 epoch and 3 nodes as the final version, deeming it the most stable and reliable. 

\section{Conclusion}

In this study, we introduce the KaLM-Embedding model, developed using a small-scale autoregressive language model. By leveraging a substantial volume of high-quality data and implementing effective training strategies, the KaLM-Embedding model achieves state-of-the-art performance in multilingual tasks at the 0.5B parameter scale. 
We have made our model open-source to facilitate access for researchers and practitioners interested in experimentation. We anticipate that our technical report will serve as a valuable resource for researchers in this field.

\section{Discussion}

Drawing from our experimental investigations and experience, we provide several discussions that may prove advantageous for future research endeavors:

Firstly, the embedding of long texts presents a promising area for further investigation. The Qwen model~\citep{DBLP:journals/corr/abs-2407-10671} we utilize employs Rotary Position Embedding (RoPE)~\citep{DBLP:journals/ijon/SuALPBL24}, which allows for the extension of context to longer sequences with relative ease. However, we have not yet adopted a single-vector approach for long text embedding due to the inherent complexity and diversity of information in long texts. Representing such texts with a single vector can lead to representation collapse or dilution~\citep{jina2023chunking,DBLP:journals/corr/abs-2410-24200}. Experimental results indicate that pure long text embedding (Dense) is less effective than BM25 (Sparse)~\citep{DBLP:journals/corr/abs-2402-03216,DBLP:conf/emnlp/ZhangZLXDTLYXHZ24,DBLP:journals/corr/abs-2410-10293}. For single-vector representations, encoding minimal and clean information is preferable~\citep{DBLP:conf/emnlp/Chen0C0MZ0024}, whereas the representation of long texts is better managed through the use of multiple vectors~\citep{DBLP:conf/acl/LuoLXZ00L24,DBLP:conf/acl/ZhangLGJD22}.

Secondly, model merging as a multi-task learning approach warrants further exploration~\citep{DBLP:conf/iclr/Jin0P023,DBLP:conf/acl/XiaoLZX24,DBLP:conf/icml/Yu0Y0L24}. In this study, we attempted to train models separately on symmetric tasks (without instruction) and asymmetric tasks (with instruction), followed by weighted averaging of the parameters. However, the separately trained models did not outperform their counterparts on their respective tasks. For example, the model trained on symmetric tasks underperformed on the STS task compared to the model trained with a mixture of all data. Additionally, the performance of the merged model deteriorated significantly, rendering it unusable. This may be attributed to a substantial gap or conflict between the two types of tasks, making direct parameter averaging infeasible~\citep{DBLP:conf/nips/YadavTCRB23}.

Lastly, the effects of different base models and pooling methods remain to be thoroughly explored. Current models employ various base models and pooling methods, yet there is a lack of rigorous comparison across different settings~\citep{DBLP:journals/corr/abs-2405-17428}. Our experiments revealed a consistent trend in the performance of different base models and pooling methods; however, the performance gap under the same training data and strategies was not significant. We posit that high-quality data is the fundamental driver for pushing the upper limits of model performance~\citep{DBLP:conf/acl/WangYHYMW24,DBLP:conf/emnlp/ZhangZLXDTLYXHZ24}, while training strategies act as catalysts that facilitate models in reaching these limits. Furthermore, the potential for innovative model architecture designs remains an exciting avenue for future research.

\bibliographystyle{plainnat}
\bibliography{reference}

\begin{thebibliography}{117}
\providecommand{\natexlab}[1]{#1}
\providecommand{\url}[1]{\texttt{#1}}
\expandafter\ifx\csname urlstyle\endcsname\relax
  \providecommand{\doi}[1]{doi: #1}\else
  \providecommand{\doi}{doi: \begingroup \urlstyle{rm}\Url}\fi

\bibitem[AI(2024)]{jina2023chunking}
Jina AI.
\newblock Still need chunking when long-context models can do it all?, 2024.
\newblock URL \url{https://jina.ai/news/still-need-chunking-when-long-context-models-can-do-it-all}.
\newblock Accessed: 2024-12-13.

\bibitem[Bonifacio et~al.(2021)Bonifacio, Campiotti, Lotufo, and Nogueira]{DBLP:journals/corr/abs-2108-13897}
Luiz~Henrique Bonifacio, Israel Campiotti, Roberto~A. Lotufo, and Rodrigo~Frassetto Nogueira.
\newblock mmarco: {A} multilingual version of {MS} {MARCO} passage ranking dataset.
\newblock \emph{CoRR}, abs/2108.13897, 2021.
\newblock URL \url{https://arxiv.org/abs/2108.13897}.

\bibitem[Boteva et~al.(2016)Boteva, Ghalandari, Sokolov, and Riezler]{DBLP:conf/ecir/BotevaGSR16}
Vera Boteva, Demian~Gholipour Ghalandari, Artem Sokolov, and Stefan Riezler.
\newblock A full-text learning to rank dataset for medical information retrieval.
\newblock In Nicola Ferro, Fabio Crestani, Marie{-}Francine Moens, Josiane Mothe, Fabrizio Silvestri, Giorgio Maria~Di Nunzio, Claudia Hauff, and Gianmaria Silvello, editors, \emph{Advances in Information Retrieval - 38th European Conference on {IR} Research, {ECIR} 2016, Padua, Italy, March 20-23, 2016. Proceedings}, volume 9626 of \emph{Lecture Notes in Computer Science}, pages 716--722. Springer, 2016.
\newblock \doi{10.1007/978-3-319-30671-1\_58}.
\newblock URL \url{https://doi.org/10.1007/978-3-319-30671-1\_58}.

\bibitem[Bowman et~al.(2015)Bowman, Angeli, Potts, and Manning]{DBLP:conf/emnlp/BowmanAPM15}
Samuel~R. Bowman, Gabor Angeli, Christopher Potts, and Christopher~D. Manning.
\newblock A large annotated corpus for learning natural language inference.
\newblock In Llu{\'{\i}}s M{\`{a}}rquez, Chris Callison{-}Burch, Jian Su, Daniele Pighin, and Yuval Marton, editors, \emph{Proceedings of the 2015 Conference on Empirical Methods in Natural Language Processing, {EMNLP} 2015, Lisbon, Portugal, September 17-21, 2015}, pages 632--642. The Association for Computational Linguistics, 2015.
\newblock \doi{10.18653/V1/D15-1075}.
\newblock URL \url{https://doi.org/10.18653/v1/d15-1075}.

\bibitem[Casanueva et~al.(2020)Casanueva, Temcinas, Gerz, Henderson, and Vulic]{DBLP:journals/corr/abs-2003-04807}
I{\~{n}}igo Casanueva, Tadas Temcinas, Daniela Gerz, Matthew Henderson, and Ivan Vulic.
\newblock Efficient intent detection with dual sentence encoders.
\newblock \emph{CoRR}, abs/2003.04807, 2020.
\newblock URL \url{https://arxiv.org/abs/2003.04807}.

\bibitem[Chan et~al.(2024)Chan, Wang, Yu, Mi, and Yu]{DBLP:journals/corr/abs-2406-20094}
Xin Chan, Xiaoyang Wang, Dian Yu, Haitao Mi, and Dong Yu.
\newblock Scaling synthetic data creation with 1,000,000,000 personas.
\newblock \emph{CoRR}, abs/2406.20094, 2024.
\newblock \doi{10.48550/ARXIV.2406.20094}.
\newblock URL \url{https://doi.org/10.48550/arXiv.2406.20094}.

\bibitem[Chen et~al.(2024{\natexlab{a}})Chen, Xiao, Zhang, Luo, Lian, and Liu]{DBLP:journals/corr/abs-2402-03216}
Jianlv Chen, Shitao Xiao, Peitian Zhang, Kun Luo, Defu Lian, and Zheng Liu.
\newblock {BGE} m3-embedding: Multi-lingual, multi-functionality, multi-granularity text embeddings through self-knowledge distillation.
\newblock \emph{CoRR}, abs/2402.03216, 2024{\natexlab{a}}.
\newblock \doi{10.48550/ARXIV.2402.03216}.
\newblock URL \url{https://doi.org/10.48550/arXiv.2402.03216}.

\bibitem[Chen et~al.(2018)Chen, Chen, Liu, Yang, Lu, and Tang]{DBLP:conf/emnlp/ChenCLYLT18}
Jing Chen, Qingcai Chen, Xin Liu, Haijun Yang, Daohe Lu, and Buzhou Tang.
\newblock The {BQ} corpus: {A} large-scale domain-specific chinese corpus for sentence semantic equivalence identification.
\newblock In Ellen Riloff, David Chiang, Julia Hockenmaier, and Jun'ichi Tsujii, editors, \emph{Proceedings of the 2018 Conference on Empirical Methods in Natural Language Processing, Brussels, Belgium, October 31 - November 4, 2018}, pages 4946--4951. Association for Computational Linguistics, 2018.
\newblock URL \url{https://aclanthology.org/D18-1536/}.

\bibitem[Chen et~al.(2024{\natexlab{b}})Chen, Wang, Chen, Yu, Ma, Zhao, Zhang, and Yu]{DBLP:conf/emnlp/Chen0C0MZ0024}
Tong Chen, Hongwei Wang, Sihao Chen, Wenhao Yu, Kaixin Ma, Xinran Zhao, Hongming Zhang, and Dong Yu.
\newblock Dense {X} retrieval: What retrieval granularity should we use?
\newblock In Yaser Al{-}Onaizan, Mohit Bansal, and Yun{-}Nung Chen, editors, \emph{Proceedings of the 2024 Conference on Empirical Methods in Natural Language Processing, {EMNLP} 2024, Miami, FL, USA, November 12-16, 2024}, pages 15159--15177. Association for Computational Linguistics, 2024{\natexlab{b}}.
\newblock URL \url{https://aclanthology.org/2024.emnlp-main.845}.

\bibitem[Ciancone et~al.(2024)Ciancone, Kerboua, Schaeffer, and Siblini]{ciancone2024mteb}
Mathieu Ciancone, Imene Kerboua, Marion Schaeffer, and Wissam Siblini.
\newblock Mteb-french: Resources for french sentence embedding evaluation and analysis.
\newblock \emph{arXiv preprint arXiv:2405.20468}, 2024.

\bibitem[Conneau et~al.(2018)Conneau, Rinott, Lample, Williams, Bowman, Schwenk, and Stoyanov]{DBLP:conf/emnlp/ConneauRLWBSS18}
Alexis Conneau, Ruty Rinott, Guillaume Lample, Adina Williams, Samuel~R. Bowman, Holger Schwenk, and Veselin Stoyanov.
\newblock {XNLI:} evaluating cross-lingual sentence representations.
\newblock In Ellen Riloff, David Chiang, Julia Hockenmaier, and Jun'ichi Tsujii, editors, \emph{Proceedings of the 2018 Conference on Empirical Methods in Natural Language Processing, Brussels, Belgium, October 31 - November 4, 2018}, pages 2475--2485. Association for Computational Linguistics, 2018.
\newblock \doi{10.18653/V1/D18-1269}.
\newblock URL \url{https://doi.org/10.18653/v1/d18-1269}.

\bibitem[Conneau et~al.(2020)Conneau, Khandelwal, Goyal, Chaudhary, Wenzek, Guzm{\'{a}}n, Grave, Ott, Zettlemoyer, and Stoyanov]{DBLP:conf/acl/ConneauKGCWGGOZ20}
Alexis Conneau, Kartikay Khandelwal, Naman Goyal, Vishrav Chaudhary, Guillaume Wenzek, Francisco Guzm{\'{a}}n, Edouard Grave, Myle Ott, Luke Zettlemoyer, and Veselin Stoyanov.
\newblock Unsupervised cross-lingual representation learning at scale.
\newblock In Dan Jurafsky, Joyce Chai, Natalie Schluter, and Joel~R. Tetreault, editors, \emph{Proceedings of the 58th Annual Meeting of the Association for Computational Linguistics, {ACL} 2020, Online, July 5-10, 2020}, pages 8440--8451. Association for Computational Linguistics, 2020.
\newblock \doi{10.18653/V1/2020.ACL-MAIN.747}.
\newblock URL \url{https://doi.org/10.18653/v1/2020.acl-main.747}.

\bibitem[Costa{-}juss{\`{a}} et~al.(2022)Costa{-}juss{\`{a}}, Cross, {\c{C}}elebi, Elbayad, Heafield, Heffernan, Kalbassi, Lam, Licht, Maillard, Sun, Wang, Wenzek, Youngblood, Akula, Barrault, Gonzalez, Hansanti, Hoffman, Jarrett, Sadagopan, Rowe, Spruit, Tran, Andrews, Ayan, Bhosale, Edunov, Fan, Gao, Goswami, Guzm{\'{a}}n, Koehn, Mourachko, Ropers, Saleem, Schwenk, and Wang]{DBLP:journals/corr/abs-2207-04672}
Marta~R. Costa{-}juss{\`{a}}, James Cross, Onur {\c{C}}elebi, Maha Elbayad, Kenneth Heafield, Kevin Heffernan, Elahe Kalbassi, Janice Lam, Daniel Licht, Jean Maillard, Anna~Y. Sun, Skyler Wang, Guillaume Wenzek, Al~Youngblood, Bapi Akula, Lo{\"{\i}}c Barrault, Gabriel~Mejia Gonzalez, Prangthip Hansanti, John Hoffman, Semarley Jarrett, Kaushik~Ram Sadagopan, Dirk Rowe, Shannon Spruit, Chau Tran, Pierre Andrews, Necip~Fazil Ayan, Shruti Bhosale, Sergey Edunov, Angela Fan, Cynthia Gao, Vedanuj Goswami, Francisco Guzm{\'{a}}n, Philipp Koehn, Alexandre Mourachko, Christophe Ropers, Safiyyah Saleem, Holger Schwenk, and Jeff Wang.
\newblock No language left behind: Scaling human-centered machine translation.
\newblock \emph{CoRR}, abs/2207.04672, 2022.
\newblock \doi{10.48550/ARXIV.2207.04672}.
\newblock URL \url{https://doi.org/10.48550/arXiv.2207.04672}.

\bibitem[Cui et~al.(2019)Cui, Liu, Che, Xiao, Chen, Ma, Wang, and Hu]{DBLP:conf/emnlp/CuiLCXCMWH19}
Yiming Cui, Ting Liu, Wanxiang Che, Li~Xiao, Zhipeng Chen, Wentao Ma, Shijin Wang, and Guoping Hu.
\newblock A span-extraction dataset for chinese machine reading comprehension.
\newblock In Kentaro Inui, Jing Jiang, Vincent Ng, and Xiaojun Wan, editors, \emph{Proceedings of the 2019 Conference on Empirical Methods in Natural Language Processing and the 9th International Joint Conference on Natural Language Processing, {EMNLP-IJCNLP} 2019, Hong Kong, China, November 3-7, 2019}, pages 5882--5888. Association for Computational Linguistics, 2019.
\newblock \doi{10.18653/V1/D19-1600}.
\newblock URL \url{https://doi.org/10.18653/v1/D19-1600}.

\bibitem[Dai et~al.(2023)Dai, Zhao, Ma, Luan, Ni, Lu, Bakalov, Guu, Hall, and Chang]{DBLP:conf/iclr/DaiZMLNLBGHC23}
Zhuyun Dai, Vincent~Y. Zhao, Ji~Ma, Yi~Luan, Jianmo Ni, Jing Lu, Anton Bakalov, Kelvin Guu, Keith~B. Hall, and Ming{-}Wei Chang.
\newblock Promptagator: Few-shot dense retrieval from 8 examples.
\newblock In \emph{The Eleventh International Conference on Learning Representations, {ICLR} 2023, Kigali, Rwanda, May 1-5, 2023}. OpenReview.net, 2023.
\newblock URL \url{https://openreview.net/forum?id=gmL46YMpu2J}.

\bibitem[DataCanary et~al.(2017)DataCanary, hilfialkaff, Jiang, Risdal, Dandekar, and tomtung]{quora-question-pairs}
DataCanary, hilfialkaff, Lili Jiang, Meg Risdal, Nikhil Dandekar, and tomtung.
\newblock Quora question pairs, 2017.
\newblock URL \url{https://kaggle.com/competitions/quora-question-pairs}.

\bibitem[Dunn et~al.(2017)Dunn, Sagun, Higgins, G{\"{u}}ney, Cirik, and Cho]{DBLP:journals/corr/DunnSHGCC17}
Matthew Dunn, Levent Sagun, Mike Higgins, V.~Ugur G{\"{u}}ney, Volkan Cirik, and Kyunghyun Cho.
\newblock Searchqa: {A} new q{\&}a dataset augmented with context from a search engine.
\newblock \emph{CoRR}, abs/1704.05179, 2017.
\newblock URL \url{http://arxiv.org/abs/1704.05179}.

\bibitem[Fader et~al.(2014)Fader, Zettlemoyer, and Etzioni]{DBLP:conf/kdd/FaderZE14}
Anthony Fader, Luke Zettlemoyer, and Oren Etzioni.
\newblock Open question answering over curated and extracted knowledge bases.
\newblock In Sofus~A. Macskassy, Claudia Perlich, Jure Leskovec, Wei Wang, and Rayid Ghani, editors, \emph{The 20th {ACM} {SIGKDD} International Conference on Knowledge Discovery and Data Mining, {KDD} '14, New York, NY, {USA} - August 24 - 27, 2014}, pages 1156--1165. {ACM}, 2014.
\newblock \doi{10.1145/2623330.2623677}.
\newblock URL \url{https://doi.org/10.1145/2623330.2623677}.

\bibitem[Fan et~al.(2019)Fan, Jernite, Perez, Grangier, Weston, and Auli]{DBLP:conf/acl/FanJPGWA19}
Angela Fan, Yacine Jernite, Ethan Perez, David Grangier, Jason Weston, and Michael Auli.
\newblock {ELI5:} long form question answering.
\newblock In Anna Korhonen, David~R. Traum, and Llu{\'{\i}}s M{\`{a}}rquez, editors, \emph{Proceedings of the 57th Conference of the Association for Computational Linguistics, {ACL} 2019, Florence, Italy, July 28- August 2, 2019, Volume 1: Long Papers}, pages 3558--3567. Association for Computational Linguistics, 2019.
\newblock \doi{10.18653/V1/P19-1346}.
\newblock URL \url{https://doi.org/10.18653/v1/p19-1346}.

\bibitem[FitzGerald et~al.(2023)FitzGerald, Hench, Peris, Mackie, Rottmann, Sanchez, Nash, Urbach, Kakarala, Singh, Ranganath, Crist, Britan, Leeuwis, T{\"{u}}r, and Natarajan]{DBLP:conf/acl/FitzGeraldHPMRS23}
Jack FitzGerald, Christopher Hench, Charith Peris, Scott Mackie, Kay Rottmann, Ana Sanchez, Aaron Nash, Liam Urbach, Vishesh Kakarala, Richa Singh, Swetha Ranganath, Laurie Crist, Misha Britan, Wouter Leeuwis, G{\"{o}}khan T{\"{u}}r, and Prem Natarajan.
\newblock {MASSIVE:} {A} 1m-example multilingual natural language understanding dataset with 51 typologically-diverse languages.
\newblock In Anna Rogers, Jordan~L. Boyd{-}Graber, and Naoaki Okazaki, editors, \emph{Proceedings of the 61st Annual Meeting of the Association for Computational Linguistics (Volume 1: Long Papers), {ACL} 2023, Toronto, Canada, July 9-14, 2023}, pages 4277--4302. Association for Computational Linguistics, 2023.
\newblock \doi{10.18653/V1/2023.ACL-LONG.235}.
\newblock URL \url{https://doi.org/10.18653/v1/2023.acl-long.235}.

\bibitem[Foundation(2024)]{wikidump}
Wikimedia Foundation.
\newblock Wikimedia downloads, 2024.
\newblock URL \url{https://dumps.wikimedia.org}.
\newblock Accessed: 2024-05-01.

\bibitem[Gao et~al.(2021)Gao, Yao, and Chen]{DBLP:conf/emnlp/GaoYC21}
Tianyu Gao, Xingcheng Yao, and Danqi Chen.
\newblock Simcse: Simple contrastive learning of sentence embeddings.
\newblock In Marie{-}Francine Moens, Xuanjing Huang, Lucia Specia, and Scott~Wen{-}tau Yih, editors, \emph{Proceedings of the 2021 Conference on Empirical Methods in Natural Language Processing, {EMNLP} 2021, Virtual Event / Punta Cana, Dominican Republic, 7-11 November, 2021}, pages 6894--6910. Association for Computational Linguistics, 2021.
\newblock \doi{10.18653/V1/2021.EMNLP-MAIN.552}.
\newblock URL \url{https://doi.org/10.18653/v1/2021.emnlp-main.552}.

\bibitem[Gao et~al.(2023)Gao, Xiong, Gao, Jia, Pan, Bi, Dai, Sun, Guo, Wang, and Wang]{DBLP:journals/corr/abs-2312-10997}
Yunfan Gao, Yun Xiong, Xinyu Gao, Kangxiang Jia, Jinliu Pan, Yuxi Bi, Yi~Dai, Jiawei Sun, Qianyu Guo, Meng Wang, and Haofen Wang.
\newblock Retrieval-augmented generation for large language models: {A} survey.
\newblock \emph{CoRR}, abs/2312.10997, 2023.
\newblock \doi{10.48550/ARXIV.2312.10997}.
\newblock URL \url{https://doi.org/10.48550/arXiv.2312.10997}.

\bibitem[Hamborg et~al.(2017)Hamborg, Meuschke, Breitinger, and Gipp]{DBLP:conf/isiwi/HamborgMBG17}
Felix Hamborg, Norman Meuschke, Corinna Breitinger, and Bela Gipp.
\newblock news-please - {A} generic news crawler and extractor.
\newblock In Maria G{\"{a}}de, Violeta Trkulja, and Vivien Petras, editors, \emph{Everything Changes, Everything Stays the Same? Understanding Information Spaces. Proceedings of the 15th International Symposium of Information Science, {ISI} 2017, Berlin, Germany, March 13-15, 2017}, volume~70 of \emph{Schriften zur Informationswissenschaft}, pages 218--223. Verlag Werner H{\"{u}}lsbusch, 2017.
\newblock \doi{10.18452/1447}.
\newblock URL \url{https://doi.org/10.18452/1447}.

\bibitem[Hasan et~al.(2021)Hasan, Bhattacharjee, Islam, Mubasshir, Li, Kang, Rahman, and Shahriyar]{DBLP:conf/acl/HasanBIMLKRS21}
Tahmid Hasan, Abhik Bhattacharjee, Md.~Saiful Islam, Kazi~Samin Mubasshir, Yuan{-}Fang Li, Yong{-}Bin Kang, M.~Sohel Rahman, and Rifat Shahriyar.
\newblock Xl-sum: Large-scale multilingual abstractive summarization for 44 languages.
\newblock In Chengqing Zong, Fei Xia, Wenjie Li, and Roberto Navigli, editors, \emph{Findings of the Association for Computational Linguistics: {ACL/IJCNLP} 2021, Online Event, August 1-6, 2021}, volume {ACL/IJCNLP} 2021 of \emph{Findings of {ACL}}, pages 4693--4703. Association for Computational Linguistics, 2021.
\newblock \doi{10.18653/V1/2021.FINDINGS-ACL.413}.
\newblock URL \url{https://doi.org/10.18653/v1/2021.findings-acl.413}.

\bibitem[He et~al.(2018)He, Liu, Liu, Lyu, Zhao, Xiao, Liu, Wang, Wu, She, Liu, Wu, and Wang]{DBLP:conf/acl/HeLLLZXLWWSLWW18}
Wei He, Kai Liu, Jing Liu, Yajuan Lyu, Shiqi Zhao, Xinyan Xiao, Yuan Liu, Yizhong Wang, Hua Wu, Qiaoqiao She, Xuan Liu, Tian Wu, and Haifeng Wang.
\newblock Dureader: a chinese machine reading comprehension dataset from real-world applications.
\newblock In Eunsol Choi, Minjoon Seo, Danqi Chen, Robin Jia, and Jonathan Berant, editors, \emph{Proceedings of the Workshop on Machine Reading for Question Answering@ACL 2018, Melbourne, Australia, July 19, 2018}, pages 37--46. Association for Computational Linguistics, 2018.
\newblock \doi{10.18653/V1/W18-2605}.
\newblock URL \url{https://aclanthology.org/W18-2605/}.

\bibitem[Heffernan et~al.(2022)Heffernan, {\c{C}}elebi, and Schwenk]{DBLP:conf/emnlp/HeffernanCS22}
Kevin Heffernan, Onur {\c{C}}elebi, and Holger Schwenk.
\newblock Bitext mining using distilled sentence representations for low-resource languages.
\newblock In Yoav Goldberg, Zornitsa Kozareva, and Yue Zhang, editors, \emph{Findings of the Association for Computational Linguistics: {EMNLP} 2022, Abu Dhabi, United Arab Emirates, December 7-11, 2022}, pages 2101--2112. Association for Computational Linguistics, 2022.
\newblock \doi{10.18653/V1/2022.FINDINGS-EMNLP.154}.
\newblock URL \url{https://doi.org/10.18653/v1/2022.findings-emnlp.154}.

\bibitem[Hou et~al.(2024)Hou, Li, He, Yan, Chen, and McAuley]{DBLP:journals/corr/abs-2403-03952}
Yupeng Hou, Jiacheng Li, Zhankui He, An~Yan, Xiusi Chen, and Julian~J. McAuley.
\newblock Bridging language and items for retrieval and recommendation.
\newblock \emph{CoRR}, abs/2403.03952, 2024.
\newblock \doi{10.48550/ARXIV.2403.03952}.
\newblock URL \url{https://doi.org/10.48550/arXiv.2403.03952}.

\bibitem[Hu et~al.(2015)Hu, Chen, and Zhu]{DBLP:conf/emnlp/HuCZ15}
Baotian Hu, Qingcai Chen, and Fangze Zhu.
\newblock {LCSTS:} {A} large scale chinese short text summarization dataset.
\newblock In Llu{\'{\i}}s M{\`{a}}rquez, Chris Callison{-}Burch, Jian Su, Daniele Pighin, and Yuval Marton, editors, \emph{Proceedings of the 2015 Conference on Empirical Methods in Natural Language Processing, {EMNLP} 2015, Lisbon, Portugal, September 17-21, 2015}, pages 1967--1972. The Association for Computational Linguistics, 2015.
\newblock \doi{10.18653/V1/D15-1229}.
\newblock URL \url{https://doi.org/10.18653/v1/d15-1229}.

\bibitem[Hu et~al.(2020)Hu, Richardson, Xu, Li, K{\"{u}}bler, and Moss]{DBLP:conf/emnlp/HuRXLKM20}
Hai Hu, Kyle Richardson, Liang Xu, Lu~Li, Sandra K{\"{u}}bler, and Lawrence~S. Moss.
\newblock {OCNLI:} original chinese natural language inference.
\newblock In Trevor Cohn, Yulan He, and Yang Liu, editors, \emph{Findings of the Association for Computational Linguistics: {EMNLP} 2020, Online Event, 16-20 November 2020}, volume {EMNLP} 2020 of \emph{Findings of {ACL}}, pages 3512--3526. Association for Computational Linguistics, 2020.
\newblock \doi{10.18653/V1/2020.FINDINGS-EMNLP.314}.
\newblock URL \url{https://doi.org/10.18653/v1/2020.findings-emnlp.314}.

\bibitem[Hu et~al.(2022)Hu, Guo, Wu, Liu, Wen, and Yu]{DBLP:conf/naacl/HuGWLWY22}
Xuming Hu, Zhijiang Guo, Guanyu Wu, Aiwei Liu, Lijie Wen, and Philip~S. Yu.
\newblock {CHEF:} {A} pilot chinese dataset for evidence-based fact-checking.
\newblock In Marine Carpuat, Marie{-}Catherine de~Marneffe, and Iv{\'{a}}n Vladimir~Meza Ru{\'{\i}}z, editors, \emph{Proceedings of the 2022 Conference of the North American Chapter of the Association for Computational Linguistics: Human Language Technologies, {NAACL} 2022, Seattle, WA, United States, July 10-15, 2022}, pages 3362--3376. Association for Computational Linguistics, 2022.
\newblock \doi{10.18653/V1/2022.NAACL-MAIN.246}.
\newblock URL \url{https://doi.org/10.18653/v1/2022.naacl-main.246}.

\bibitem[Huang et~al.(2024)Huang, Hu, Jing, Gao, and Wu]{DBLP:journals/corr/abs-2405-06932}
Junqin Huang, Zhongjie Hu, Zihao Jing, Mengya Gao, and Yichao Wu.
\newblock Piccolo2: General text embedding with multi-task hybrid loss training.
\newblock \emph{CoRR}, abs/2405.06932, 2024.
\newblock \doi{10.48550/ARXIV.2405.06932}.
\newblock URL \url{https://doi.org/10.48550/arXiv.2405.06932}.

\bibitem[Huang and Huang(2024)]{DBLP:journals/corr/abs-2404-10981}
Yizheng Huang and Jimmy Huang.
\newblock A survey on retrieval-augmented text generation for large language models.
\newblock \emph{CoRR}, abs/2404.10981, 2024.
\newblock \doi{10.48550/ARXIV.2404.10981}.
\newblock URL \url{https://doi.org/10.48550/arXiv.2404.10981}.

\bibitem[Husain et~al.(2019)Husain, Wu, Gazit, Allamanis, and Brockschmidt]{DBLP:journals/corr/abs-1909-09436}
Hamel Husain, Ho{-}Hsiang Wu, Tiferet Gazit, Miltiadis Allamanis, and Marc Brockschmidt.
\newblock Codesearchnet challenge: Evaluating the state of semantic code search.
\newblock \emph{CoRR}, abs/1909.09436, 2019.
\newblock URL \url{http://arxiv.org/abs/1909.09436}.

\bibitem[Jin et~al.(2019)Jin, Dhingra, Liu, Cohen, and Lu]{DBLP:conf/emnlp/JinDLCL19}
Qiao Jin, Bhuwan Dhingra, Zhengping Liu, William~W. Cohen, and Xinghua Lu.
\newblock Pubmedqa: {A} dataset for biomedical research question answering.
\newblock In Kentaro Inui, Jing Jiang, Vincent Ng, and Xiaojun Wan, editors, \emph{Proceedings of the 2019 Conference on Empirical Methods in Natural Language Processing and the 9th International Joint Conference on Natural Language Processing, {EMNLP-IJCNLP} 2019, Hong Kong, China, November 3-7, 2019}, pages 2567--2577. Association for Computational Linguistics, 2019.
\newblock \doi{10.18653/V1/D19-1259}.
\newblock URL \url{https://doi.org/10.18653/v1/D19-1259}.

\bibitem[Jin et~al.(2023)Jin, Ren, Preotiuc{-}Pietro, and Cheng]{DBLP:conf/iclr/Jin0P023}
Xisen Jin, Xiang Ren, Daniel Preotiuc{-}Pietro, and Pengxiang Cheng.
\newblock Dataless knowledge fusion by merging weights of language models.
\newblock In \emph{The Eleventh International Conference on Learning Representations, {ICLR} 2023, Kigali, Rwanda, May 1-5, 2023}. OpenReview.net, 2023.
\newblock URL \url{https://openreview.net/forum?id=FCnohuR6AnM}.

\bibitem[Joshi et~al.(2017)Joshi, Choi, Weld, and Zettlemoyer]{DBLP:conf/acl/JoshiCWZ17}
Mandar Joshi, Eunsol Choi, Daniel~S. Weld, and Luke Zettlemoyer.
\newblock Triviaqa: {A} large scale distantly supervised challenge dataset for reading comprehension.
\newblock In Regina Barzilay and Min{-}Yen Kan, editors, \emph{Proceedings of the 55th Annual Meeting of the Association for Computational Linguistics, {ACL} 2017, Vancouver, Canada, July 30 - August 4, Volume 1: Long Papers}, pages 1601--1611. Association for Computational Linguistics, 2017.
\newblock \doi{10.18653/V1/P17-1147}.
\newblock URL \url{https://doi.org/10.18653/v1/P17-1147}.

\bibitem[Khashabi et~al.(2021)Khashabi, Ng, Khot, Sabharwal, Hajishirzi, and Callison{-}Burch]{DBLP:conf/emnlp/KhashabiNKSHC21}
Daniel Khashabi, Amos Ng, Tushar Khot, Ashish Sabharwal, Hannaneh Hajishirzi, and Chris Callison{-}Burch.
\newblock Gooaq: Open question answering with diverse answer types.
\newblock In Marie{-}Francine Moens, Xuanjing Huang, Lucia Specia, and Scott~Wen{-}tau Yih, editors, \emph{Findings of the Association for Computational Linguistics: {EMNLP} 2021, Virtual Event / Punta Cana, Dominican Republic, 16-20 November, 2021}, pages 421--433. Association for Computational Linguistics, 2021.
\newblock \doi{10.18653/V1/2021.FINDINGS-EMNLP.38}.
\newblock URL \url{https://doi.org/10.18653/v1/2021.findings-emnlp.38}.

\bibitem[Koreeda and Manning(2021)]{DBLP:conf/emnlp/KoreedaM21}
Yuta Koreeda and Christopher~D. Manning.
\newblock Contractnli: {A} dataset for document-level natural language inference for contracts.
\newblock In Marie{-}Francine Moens, Xuanjing Huang, Lucia Specia, and Scott~Wen{-}tau Yih, editors, \emph{Findings of the Association for Computational Linguistics: {EMNLP} 2021, Virtual Event / Punta Cana, Dominican Republic, 16-20 November, 2021}, pages 1907--1919. Association for Computational Linguistics, 2021.
\newblock \doi{10.18653/V1/2021.FINDINGS-EMNLP.164}.
\newblock URL \url{https://doi.org/10.18653/v1/2021.findings-emnlp.164}.

\bibitem[Kusupati et~al.(2022)Kusupati, Bhatt, Rege, Wallingford, Sinha, Ramanujan, Howard{-}Snyder, Chen, Kakade, Jain, and Farhadi]{DBLP:conf/nips/KusupatiBRWSRHC22}
Aditya Kusupati, Gantavya Bhatt, Aniket Rege, Matthew Wallingford, Aditya Sinha, Vivek Ramanujan, William Howard{-}Snyder, Kaifeng Chen, Sham~M. Kakade, Prateek Jain, and Ali Farhadi.
\newblock Matryoshka representation learning.
\newblock In Sanmi Koyejo, S.~Mohamed, A.~Agarwal, Danielle Belgrave, K.~Cho, and A.~Oh, editors, \emph{Advances in Neural Information Processing Systems 35: Annual Conference on Neural Information Processing Systems 2022, NeurIPS 2022, New Orleans, LA, USA, November 28 - December 9, 2022}, 2022.
\newblock URL \url{http://papers.nips.cc/paper\_files/paper/2022/hash/c32319f4868da7613d78af9993100e42-Abstract-Conference.html}.

\bibitem[Kwiatkowski et~al.(2019)Kwiatkowski, Palomaki, Redfield, Collins, Parikh, Alberti, Epstein, Polosukhin, Devlin, Lee, Toutanova, Jones, Kelcey, Chang, Dai, Uszkoreit, Le, and Petrov]{DBLP:journals/tacl/KwiatkowskiPRCP19}
Tom Kwiatkowski, Jennimaria Palomaki, Olivia Redfield, Michael Collins, Ankur~P. Parikh, Chris Alberti, Danielle Epstein, Illia Polosukhin, Jacob Devlin, Kenton Lee, Kristina Toutanova, Llion Jones, Matthew Kelcey, Ming{-}Wei Chang, Andrew~M. Dai, Jakob Uszkoreit, Quoc Le, and Slav Petrov.
\newblock Natural questions: a benchmark for question answering research.
\newblock \emph{Trans. Assoc. Comput. Linguistics}, 7:\penalty0 452--466, 2019.
\newblock \doi{10.1162/TACL\_A\_00276}.
\newblock URL \url{https://doi.org/10.1162/tacl\_a\_00276}.

\bibitem[Lee et~al.(2024)Lee, Roy, Xu, Raiman, Shoeybi, Catanzaro, and Ping]{DBLP:journals/corr/abs-2405-17428}
Chankyu Lee, Rajarshi Roy, Mengyao Xu, Jonathan Raiman, Mohammad Shoeybi, Bryan Catanzaro, and Wei Ping.
\newblock Nv-embed: Improved techniques for training llms as generalist embedding models.
\newblock \emph{CoRR}, abs/2405.17428, 2024.
\newblock \doi{10.48550/ARXIV.2405.17428}.
\newblock URL \url{https://doi.org/10.48550/arXiv.2405.17428}.

\bibitem[Lewis et~al.(2021)Lewis, Wu, Liu, Minervini, K{\"{u}}ttler, Piktus, Stenetorp, and Riedel]{DBLP:journals/tacl/LewisWLMKPSR21}
Patrick S.~H. Lewis, Yuxiang Wu, Linqing Liu, Pasquale Minervini, Heinrich K{\"{u}}ttler, Aleksandra Piktus, Pontus Stenetorp, and Sebastian Riedel.
\newblock {PAQ:} 65 million probably-asked questions and what you can do with them.
\newblock \emph{Trans. Assoc. Comput. Linguistics}, 9:\penalty0 1098--1115, 2021.
\newblock \doi{10.1162/TACL\_A\_00415}.
\newblock URL \url{https://doi.org/10.1162/tacl\_a\_00415}.

\bibitem[Li et~al.(2021)Li, Arora, Chen, Gupta, Gupta, and Mehdad]{DBLP:conf/eacl/LiACGGM21}
Haoran Li, Abhinav Arora, Shuohui Chen, Anchit Gupta, Sonal Gupta, and Yashar Mehdad.
\newblock {MTOP:} {A} comprehensive multilingual task-oriented semantic parsing benchmark.
\newblock In Paola Merlo, J{\"{o}}rg Tiedemann, and Reut Tsarfaty, editors, \emph{Proceedings of the 16th Conference of the European Chapter of the Association for Computational Linguistics: Main Volume, {EACL} 2021, Online, April 19 - 23, 2021}, pages 2950--2962. Association for Computational Linguistics, 2021.
\newblock \doi{10.18653/V1/2021.EACL-MAIN.257}.
\newblock URL \url{https://doi.org/10.18653/v1/2021.eacl-main.257}.

\bibitem[Li et~al.(2022)Li, Zhang, Zhao, Shen, Liu, Mao, and Zhang]{DBLP:conf/coling/LiZ0S0MZ22}
Yudong Li, Yuqing Zhang, Zhe Zhao, Linlin Shen, Weijie Liu, Weiquan Mao, and Hui Zhang.
\newblock {CSL:} {A} large-scale chinese scientific literature dataset.
\newblock In Nicoletta Calzolari, Chu{-}Ren Huang, Hansaem Kim, James Pustejovsky, Leo Wanner, Key{-}Sun Choi, Pum{-}Mo Ryu, Hsin{-}Hsi Chen, Lucia Donatelli, Heng Ji, Sadao Kurohashi, Patrizia Paggio, Nianwen Xue, Seokhwan Kim, Younggyun Hahm, Zhong He, Tony~Kyungil Lee, Enrico Santus, Francis Bond, and Seung{-}Hoon Na, editors, \emph{Proceedings of the 29th International Conference on Computational Linguistics, {COLING} 2022, Gyeongju, Republic of Korea, October 12-17, 2022}, pages 3917--3923. International Committee on Computational Linguistics, 2022.
\newblock URL \url{https://aclanthology.org/2022.coling-1.344}.

\bibitem[Li et~al.(2023)Li, Zhang, Zhang, Long, Xie, and Zhang]{DBLP:journals/corr/abs-2308-03281}
Zehan Li, Xin Zhang, Yanzhao Zhang, Dingkun Long, Pengjun Xie, and Meishan Zhang.
\newblock Towards general text embeddings with multi-stage contrastive learning.
\newblock \emph{CoRR}, abs/2308.03281, 2023.
\newblock \doi{10.48550/ARXIV.2308.03281}.
\newblock URL \url{https://doi.org/10.48550/arXiv.2308.03281}.

\bibitem[Liu et~al.(2023)Liu, Liao, Meng, and Wang]{LAWGPT-zh}
Hongcheng Liu, Yusheng Liao, Yutong Meng, and Yuhao Wang.
\newblock Xiezhi: Chinese law large language model.
\newblock https://github.com/LiuHC0428/LAW\_GPT, 2023.

\bibitem[Liu et~al.(2018)Liu, Chen, Deng, Zeng, Chen, Li, and Tang]{DBLP:conf/coling/LiuCDZCLT18}
Xin Liu, Qingcai Chen, Chong Deng, Huajun Zeng, Jing Chen, Dongfang Li, and Buzhou Tang.
\newblock {LCQMC:} {A} large-scale chinese question matching corpus.
\newblock In Emily~M. Bender, Leon Derczynski, and Pierre Isabelle, editors, \emph{Proceedings of the 27th International Conference on Computational Linguistics, {COLING} 2018, Santa Fe, New Mexico, USA, August 20-26, 2018}, pages 1952--1962. Association for Computational Linguistics, 2018.
\newblock URL \url{https://aclanthology.org/C18-1166/}.

\bibitem[Long et~al.(2022)Long, Gao, Zou, Xu, Xie, Guo, Xu, Jiang, Xing, and Yang]{DBLP:conf/sigir/LongGZXXGXJXY22}
Dingkun Long, Qiong Gao, Kuan Zou, Guangwei Xu, Pengjun Xie, Ruijie Guo, Jian Xu, Guanjun Jiang, Luxi Xing, and Ping Yang.
\newblock Multi-cpr: {A} multi domain chinese dataset for passage retrieval.
\newblock In Enrique Amig{\'{o}}, Pablo Castells, Julio Gonzalo, Ben Carterette, J.~Shane Culpepper, and Gabriella Kazai, editors, \emph{{SIGIR} '22: The 45th International {ACM} {SIGIR} Conference on Research and Development in Information Retrieval, Madrid, Spain, July 11 - 15, 2022}, pages 3046--3056. {ACM}, 2022.
\newblock \doi{10.1145/3477495.3531736}.
\newblock URL \url{https://doi.org/10.1145/3477495.3531736}.

\bibitem[Luo et~al.(2024)Luo, Liu, Xiao, Zhou, Chen, Zhao, and Liu]{DBLP:conf/acl/LuoLXZ00L24}
Kun Luo, Zheng Liu, Shitao Xiao, Tong Zhou, Yubo Chen, Jun Zhao, and Kang Liu.
\newblock Landmark embedding: {A} chunking-free embedding method for retrieval augmented long-context large language models.
\newblock In Lun{-}Wei Ku, Andre Martins, and Vivek Srikumar, editors, \emph{Proceedings of the 62nd Annual Meeting of the Association for Computational Linguistics (Volume 1: Long Papers), {ACL} 2024, Bangkok, Thailand, August 11-16, 2024}, pages 3268--3281. Association for Computational Linguistics, 2024.
\newblock \doi{10.18653/V1/2024.ACL-LONG.180}.
\newblock URL \url{https://doi.org/10.18653/v1/2024.acl-long.180}.

\bibitem[Maas et~al.(2011)Maas, Daly, Pham, Huang, Ng, and Potts]{DBLP:conf/acl/MaasDPHNP11}
Andrew~L. Maas, Raymond~E. Daly, Peter~T. Pham, Dan Huang, Andrew~Y. Ng, and Christopher Potts.
\newblock Learning word vectors for sentiment analysis.
\newblock In Dekang Lin, Yuji Matsumoto, and Rada Mihalcea, editors, \emph{The 49th Annual Meeting of the Association for Computational Linguistics: Human Language Technologies, Proceedings of the Conference, 19-24 June, 2011, Portland, Oregon, {USA}}, pages 142--150. The Association for Computer Linguistics, 2011.
\newblock URL \url{https://aclanthology.org/P11-1015/}.

\bibitem[Maia et~al.(2018)Maia, Handschuh, Freitas, Davis, McDermott, Zarrouk, and Balahur]{DBLP:conf/www/MaiaHFDMZB18}
Macedo Maia, Siegfried Handschuh, Andr{\'{e}} Freitas, Brian Davis, Ross McDermott, Manel Zarrouk, and Alexandra Balahur.
\newblock Www'18 open challenge: Financial opinion mining and question answering.
\newblock In Pierre{-}Antoine Champin, Fabien Gandon, Mounia Lalmas, and Panagiotis~G. Ipeirotis, editors, \emph{Companion of the The Web Conference 2018 on The Web Conference 2018, {WWW} 2018, Lyon , France, April 23-27, 2018}, pages 1941--1942. {ACM}, 2018.
\newblock \doi{10.1145/3184558.3192301}.
\newblock URL \url{https://doi.org/10.1145/3184558.3192301}.

\bibitem[Malaviya et~al.(2024)Malaviya, Lee, Chen, Sieber, Yatskar, and Roth]{DBLP:conf/naacl/MalaviyaLCSYR24}
Chaitanya Malaviya, Subin Lee, Sihao Chen, Elizabeth Sieber, Mark Yatskar, and Dan Roth.
\newblock Expertqa: Expert-curated questions and attributed answers.
\newblock In Kevin Duh, Helena G{\'{o}}mez{-}Adorno, and Steven Bethard, editors, \emph{Proceedings of the 2024 Conference of the North American Chapter of the Association for Computational Linguistics: Human Language Technologies (Volume 1: Long Papers), {NAACL} 2024, Mexico City, Mexico, June 16-21, 2024}, pages 3025--3045. Association for Computational Linguistics, 2024.
\newblock \doi{10.18653/V1/2024.NAACL-LONG.167}.
\newblock URL \url{https://doi.org/10.18653/v1/2024.naacl-long.167}.

\bibitem[McAuley and Leskovec(2013)]{DBLP:conf/recsys/McAuleyL13}
Julian~J. McAuley and Jure Leskovec.
\newblock Hidden factors and hidden topics: understanding rating dimensions with review text.
\newblock In Qiang Yang, Irwin King, Qing Li, Pearl Pu, and George Karypis, editors, \emph{Seventh {ACM} Conference on Recommender Systems, RecSys '13, Hong Kong, China, October 12-16, 2013}, pages 165--172. {ACM}, 2013.
\newblock \doi{10.1145/2507157.2507163}.
\newblock URL \url{https://doi.org/10.1145/2507157.2507163}.

\bibitem[Meng et~al.(2024)Meng, Liu, Joty, Xiong, Zhou, and Yavuz]{SFRAIResearch2024}
Rui Meng, Ye~Liu, Shafiq~Rayhan Joty, Caiming Xiong, Yingbo Zhou, and Semih Yavuz.
\newblock Sfr-embedding-mistral:enhance text retrieval with transfer learning.
\newblock Salesforce AI Research Blog, 2024.
\newblock URL \url{https://blog.salesforceairesearch.com/sfr-embedded-mistral/}.

\bibitem[Mollanorozy et~al.(2023)Mollanorozy, Tanti, and Nissim]{mollanorozy2023cross}
Sepideh Mollanorozy, Marc Tanti, and Malvina Nissim.
\newblock Cross-lingual transfer learning with persian.
\newblock In \emph{Proceedings of the 5th Workshop on Research in Computational Linguistic Typology and Multilingual NLP}, pages 89--95, 2023.

\bibitem[Muennighoff et~al.(2023{\natexlab{a}})Muennighoff, Tazi, Magne, and Reimers]{DBLP:conf/eacl/MuennighoffTMR23}
Niklas Muennighoff, Nouamane Tazi, Lo{\"{\i}}c Magne, and Nils Reimers.
\newblock {MTEB:} massive text embedding benchmark.
\newblock In Andreas Vlachos and Isabelle Augenstein, editors, \emph{Proceedings of the 17th Conference of the European Chapter of the Association for Computational Linguistics, {EACL} 2023, Dubrovnik, Croatia, May 2-6, 2023}, pages 2006--2029. Association for Computational Linguistics, 2023{\natexlab{a}}.
\newblock \doi{10.18653/V1/2023.EACL-MAIN.148}.
\newblock URL \url{https://doi.org/10.18653/v1/2023.eacl-main.148}.

\bibitem[Muennighoff et~al.(2023{\natexlab{b}})Muennighoff, Wang, Sutawika, Roberts, Biderman, Scao, Bari, Shen, Yong, Schoelkopf, Tang, Radev, Aji, Almubarak, Albanie, Alyafeai, Webson, Raff, and Raffel]{DBLP:conf/acl/MuennighoffWSRB23}
Niklas Muennighoff, Thomas Wang, Lintang Sutawika, Adam Roberts, Stella Biderman, Teven~Le Scao, M.~Saiful Bari, Sheng Shen, Zheng~Xin Yong, Hailey Schoelkopf, Xiangru Tang, Dragomir Radev, Alham~Fikri Aji, Khalid Almubarak, Samuel Albanie, Zaid Alyafeai, Albert Webson, Edward Raff, and Colin Raffel.
\newblock Crosslingual generalization through multitask finetuning.
\newblock In Anna Rogers, Jordan~L. Boyd{-}Graber, and Naoaki Okazaki, editors, \emph{Proceedings of the 61st Annual Meeting of the Association for Computational Linguistics (Volume 1: Long Papers), {ACL} 2023, Toronto, Canada, July 9-14, 2023}, pages 15991--16111. Association for Computational Linguistics, 2023{\natexlab{b}}.
\newblock \doi{10.18653/V1/2023.ACL-LONG.891}.
\newblock URL \url{https://doi.org/10.18653/v1/2023.acl-long.891}.

\bibitem[Muennighoff et~al.(2024)Muennighoff, Su, Wang, Yang, Wei, Yu, Singh, and Kiela]{DBLP:journals/corr/abs-2402-09906}
Niklas Muennighoff, Hongjin Su, Liang Wang, Nan Yang, Furu Wei, Tao Yu, Amanpreet Singh, and Douwe Kiela.
\newblock Generative representational instruction tuning.
\newblock \emph{CoRR}, abs/2402.09906, 2024.
\newblock \doi{10.48550/ARXIV.2402.09906}.
\newblock URL \url{https://doi.org/10.48550/arXiv.2402.09906}.

\bibitem[Mukherjee et~al.(2023)Mukherjee, Mitra, Jawahar, Agarwal, Palangi, and Awadallah]{DBLP:journals/corr/abs-2306-02707}
Subhabrata Mukherjee, Arindam Mitra, Ganesh Jawahar, Sahaj Agarwal, Hamid Palangi, and Ahmed Awadallah.
\newblock Orca: Progressive learning from complex explanation traces of {GPT-4}.
\newblock \emph{CoRR}, abs/2306.02707, 2023.
\newblock \doi{10.48550/ARXIV.2306.02707}.
\newblock URL \url{https://doi.org/10.48550/arXiv.2306.02707}.

\bibitem[Nakano et~al.(2021)Nakano, Hilton, Balaji, Wu, Ouyang, Kim, Hesse, Jain, Kosaraju, Saunders, Jiang, Cobbe, Eloundou, Krueger, Button, Knight, Chess, and Schulman]{DBLP:journals/corr/abs-2112-09332}
Reiichiro Nakano, Jacob Hilton, Suchir Balaji, Jeff Wu, Long Ouyang, Christina Kim, Christopher Hesse, Shantanu Jain, Vineet Kosaraju, William Saunders, Xu~Jiang, Karl Cobbe, Tyna Eloundou, Gretchen Krueger, Kevin Button, Matthew Knight, Benjamin Chess, and John Schulman.
\newblock Webgpt: Browser-assisted question-answering with human feedback.
\newblock \emph{CoRR}, abs/2112.09332, 2021.
\newblock URL \url{https://arxiv.org/abs/2112.09332}.

\bibitem[Nguyen et~al.(2016)Nguyen, Rosenberg, Song, Gao, Tiwary, Majumder, and Deng]{DBLP:conf/nips/NguyenRSGTMD16}
Tri Nguyen, Mir Rosenberg, Xia Song, Jianfeng Gao, Saurabh Tiwary, Rangan Majumder, and Li~Deng.
\newblock {MS} {MARCO:} {A} human generated machine reading comprehension dataset.
\newblock In Tarek~Richard Besold, Antoine Bordes, Artur~S. d'Avila Garcez, and Greg Wayne, editors, \emph{Proceedings of the Workshop on Cognitive Computation: Integrating neural and symbolic approaches 2016 co-located with the 30th Annual Conference on Neural Information Processing Systems {(NIPS} 2016), Barcelona, Spain, December 9, 2016}, volume 1773 of \emph{{CEUR} Workshop Proceedings}. CEUR-WS.org, 2016.
\newblock URL \url{https://ceur-ws.org/Vol-1773/CoCoNIPS\_2016\_paper9.pdf}.

\bibitem[Ni et~al.(2019)Ni, Li, and McAuley]{DBLP:conf/emnlp/NiLM19}
Jianmo Ni, Jiacheng Li, and Julian~J. McAuley.
\newblock Justifying recommendations using distantly-labeled reviews and fine-grained aspects.
\newblock In Kentaro Inui, Jing Jiang, Vincent Ng, and Xiaojun Wan, editors, \emph{Proceedings of the 2019 Conference on Empirical Methods in Natural Language Processing and the 9th International Joint Conference on Natural Language Processing, {EMNLP-IJCNLP} 2019, Hong Kong, China, November 3-7, 2019}, pages 188--197. Association for Computational Linguistics, 2019.
\newblock \doi{10.18653/V1/D19-1018}.
\newblock URL \url{https://doi.org/10.18653/v1/D19-1018}.

\bibitem[O'Neill et~al.(2021)O'Neill, Rozenshtein, Kiryo, Kubota, and Bollegala]{DBLP:conf/emnlp/ONeillRKKB21}
James O'Neill, Polina Rozenshtein, Ryuichi Kiryo, Motoko Kubota, and Danushka Bollegala.
\newblock I wish {I} would have loved this one, but {I} didn't - {A} multilingual dataset for counterfactual detection in product review.
\newblock In Marie{-}Francine Moens, Xuanjing Huang, Lucia Specia, and Scott~Wen{-}tau Yih, editors, \emph{Proceedings of the 2021 Conference on Empirical Methods in Natural Language Processing, {EMNLP} 2021, Virtual Event / Punta Cana, Dominican Republic, 7-11 November, 2021}, pages 7092--7108. Association for Computational Linguistics, 2021.
\newblock \doi{10.18653/V1/2021.EMNLP-MAIN.568}.
\newblock URL \url{https://doi.org/10.18653/v1/2021.emnlp-main.568}.

\bibitem[Poswiata et~al.(2024)Poswiata, Dadas, and Perelkiewicz]{DBLP:journals/corr/abs-2405-10138}
Rafal Poswiata, Slawomir Dadas, and Michal Perelkiewicz.
\newblock {PL-MTEB:} polish massive text embedding benchmark.
\newblock \emph{CoRR}, abs/2405.10138, 2024.
\newblock \doi{10.48550/ARXIV.2405.10138}.
\newblock URL \url{https://doi.org/10.48550/arXiv.2405.10138}.

\bibitem[Qin et~al.(2023)Qin, Cai, Jin, Yan, Liang, Zhu, Lin, Han, Ding, Wang, Xie, Qi, Liu, Sun, and Zhou]{DBLP:conf/acl/QinCJYLZLHDWXQL23}
Yujia Qin, Zihan Cai, Dian Jin, Lan Yan, Shihao Liang, Kunlun Zhu, Yankai Lin, Xu~Han, Ning Ding, Huadong Wang, Ruobing Xie, Fanchao Qi, Zhiyuan Liu, Maosong Sun, and Jie Zhou.
\newblock Webcpm: Interactive web search for chinese long-form question answering.
\newblock In Anna Rogers, Jordan~L. Boyd{-}Graber, and Naoaki Okazaki, editors, \emph{Proceedings of the 61st Annual Meeting of the Association for Computational Linguistics (Volume 1: Long Papers), {ACL} 2023, Toronto, Canada, July 9-14, 2023}, pages 8968--8988. Association for Computational Linguistics, 2023.
\newblock \doi{10.18653/V1/2023.ACL-LONG.499}.
\newblock URL \url{https://doi.org/10.18653/v1/2023.acl-long.499}.

\bibitem[Rajpurkar et~al.(2016)Rajpurkar, Zhang, Lopyrev, and Liang]{DBLP:conf/emnlp/RajpurkarZLL16}
Pranav Rajpurkar, Jian Zhang, Konstantin Lopyrev, and Percy Liang.
\newblock Squad: 100, 000+ questions for machine comprehension of text.
\newblock In Jian Su, Xavier Carreras, and Kevin Duh, editors, \emph{Proceedings of the 2016 Conference on Empirical Methods in Natural Language Processing, {EMNLP} 2016, Austin, Texas, USA, November 1-4, 2016}, pages 2383--2392. The Association for Computational Linguistics, 2016.
\newblock \doi{10.18653/V1/D16-1264}.
\newblock URL \url{https://doi.org/10.18653/v1/d16-1264}.

\bibitem[Rajpurkar et~al.(2018)Rajpurkar, Jia, and Liang]{DBLP:conf/acl/RajpurkarJL18}
Pranav Rajpurkar, Robin Jia, and Percy Liang.
\newblock Know what you don't know: Unanswerable questions for squad.
\newblock In Iryna Gurevych and Yusuke Miyao, editors, \emph{Proceedings of the 56th Annual Meeting of the Association for Computational Linguistics, {ACL} 2018, Melbourne, Australia, July 15-20, 2018, Volume 2: Short Papers}, pages 784--789. Association for Computational Linguistics, 2018.
\newblock \doi{10.18653/V1/P18-2124}.
\newblock URL \url{https://aclanthology.org/P18-2124/}.

\bibitem[Reddy et~al.(2022)Reddy, M{\`{a}}rquez, Valero, Rao, Zaragoza, Bandyopadhyay, Biswas, Xing, and Subbian]{DBLP:journals/corr/abs-2206-06588}
Chandan~K. Reddy, Llu{\'{\i}}s M{\`{a}}rquez, Fran Valero, Nikhil Rao, Hugo Zaragoza, Sambaran Bandyopadhyay, Arnab Biswas, Anlu Xing, and Karthik Subbian.
\newblock Shopping queries dataset: {A} large-scale {ESCI} benchmark for improving product search.
\newblock \emph{CoRR}, abs/2206.06588, 2022.
\newblock \doi{10.48550/ARXIV.2206.06588}.
\newblock URL \url{https://doi.org/10.48550/arXiv.2206.06588}.

\bibitem[Reimers and Gurevych(2019)]{DBLP:conf/emnlp/ReimersG19}
Nils Reimers and Iryna Gurevych.
\newblock Sentence-bert: Sentence embeddings using siamese bert-networks.
\newblock In Kentaro Inui, Jing Jiang, Vincent Ng, and Xiaojun Wan, editors, \emph{Proceedings of the 2019 Conference on Empirical Methods in Natural Language Processing and the 9th International Joint Conference on Natural Language Processing, {EMNLP-IJCNLP} 2019, Hong Kong, China, November 3-7, 2019}, pages 3980--3990. Association for Computational Linguistics, 2019.
\newblock \doi{10.18653/V1/D19-1410}.
\newblock URL \url{https://doi.org/10.18653/v1/D19-1410}.

\bibitem[Saravia et~al.(2018)Saravia, Liu, Huang, Wu, and Chen]{DBLP:conf/emnlp/SaraviaLHWC18}
Elvis Saravia, Hsien{-}Chi~Toby Liu, Yen{-}Hao Huang, Junlin Wu, and Yi{-}Shin Chen.
\newblock {CARER:} contextualized affect representations for emotion recognition.
\newblock In Ellen Riloff, David Chiang, Julia Hockenmaier, and Jun'ichi Tsujii, editors, \emph{Proceedings of the 2018 Conference on Empirical Methods in Natural Language Processing, Brussels, Belgium, October 31 - November 4, 2018}, pages 3687--3697. Association for Computational Linguistics, 2018.
\newblock \doi{10.18653/V1/D18-1404}.
\newblock URL \url{https://doi.org/10.18653/v1/d18-1404}.

\bibitem[Schwenk et~al.(2021)Schwenk, Wenzek, Edunov, Grave, Joulin, and Fan]{DBLP:conf/acl/SchwenkWEGJF20}
Holger Schwenk, Guillaume Wenzek, Sergey Edunov, Edouard Grave, Armand Joulin, and Angela Fan.
\newblock Ccmatrix: Mining billions of high-quality parallel sentences on the web.
\newblock In Chengqing Zong, Fei Xia, Wenjie Li, and Roberto Navigli, editors, \emph{Proceedings of the 59th Annual Meeting of the Association for Computational Linguistics and the 11th International Joint Conference on Natural Language Processing, {ACL/IJCNLP} 2021, (Volume 1: Long Papers), Virtual Event, August 1-6, 2021}, pages 6490--6500. Association for Computational Linguistics, 2021.
\newblock \doi{10.18653/V1/2021.ACL-LONG.507}.
\newblock URL \url{https://doi.org/10.18653/v1/2021.acl-long.507}.

\bibitem[Setty et~al.(2024)Setty, Jijo, Chung, and Vidra]{DBLP:journals/corr/abs-2404-07221}
Spurthi Setty, Katherine Jijo, Eden Chung, and Natan Vidra.
\newblock Improving retrieval for {RAG} based question answering models on financial documents.
\newblock \emph{CoRR}, abs/2404.07221, 2024.
\newblock \doi{10.48550/ARXIV.2404.07221}.
\newblock URL \url{https://doi.org/10.48550/arXiv.2404.07221}.

\bibitem[Shao et~al.(2018)Shao, Liu, Lai, Tseng, and Tsai]{DBLP:journals/corr/abs-1806-00920}
Chih{-}Chieh Shao, Trois Liu, Yuting Lai, Yiying Tseng, and Sam Tsai.
\newblock {DRCD:} a chinese machine reading comprehension dataset.
\newblock \emph{CoRR}, abs/1806.00920, 2018.
\newblock URL \url{http://arxiv.org/abs/1806.00920}.

\bibitem[Shao et~al.(2019)Shao, Huang, Wen, Xu, and Zhu]{DBLP:conf/emnlp/ShaoHWXZ19}
Zhihong Shao, Minlie Huang, Jiangtao Wen, Wenfei Xu, and Xiaoyan Zhu.
\newblock Long and diverse text generation with planning-based hierarchical variational model.
\newblock In Kentaro Inui, Jing Jiang, Vincent Ng, and Xiaojun Wan, editors, \emph{Proceedings of the 2019 Conference on Empirical Methods in Natural Language Processing and the 9th International Joint Conference on Natural Language Processing, {EMNLP-IJCNLP} 2019, Hong Kong, China, November 3-7, 2019}, pages 3255--3266. Association for Computational Linguistics, 2019.
\newblock \doi{10.18653/V1/D19-1321}.
\newblock URL \url{https://doi.org/10.18653/v1/D19-1321}.

\bibitem[Singh et~al.(2024)Singh, Vargus, D'souza, Karlsson, Mahendiran, Ko, Shandilya, Patel, Mataciunas, O'Mahony, Zhang, Hettiarachchi, Wilson, Machado, Moura, Krzeminski, Fadaei, Erg{\"{u}}n, Okoh, Alaagib, Mudannayake, Alyafeai, Chien, Ruder, Guthikonda, Alghamdi, Gehrmann, Muennighoff, Bartolo, Kreutzer, {\"{U}}st{\"{u}}n, Fadaee, and Hooker]{DBLP:conf/acl/SinghVD0MKSPMOZ24}
Shivalika Singh, Freddie Vargus, Daniel D'souza, B{\"{o}}rje Karlsson, Abinaya Mahendiran, Wei{-}Yin Ko, Herumb Shandilya, Jay Patel, Deividas Mataciunas, Laura O'Mahony, Mike Zhang, Ramith Hettiarachchi, Joseph Wilson, Marina Machado, Luisa~Souza Moura, Dominik Krzeminski, Hakimeh Fadaei, Irem Erg{\"{u}}n, Ifeoma Okoh, Aisha Alaagib, Oshan Mudannayake, Zaid Alyafeai, Minh~Vu Chien, Sebastian Ruder, Surya Guthikonda, Emad~A. Alghamdi, Sebastian Gehrmann, Niklas Muennighoff, Max Bartolo, Julia Kreutzer, Ahmet {\"{U}}st{\"{u}}n, Marzieh Fadaee, and Sara Hooker.
\newblock Aya dataset: An open-access collection for multilingual instruction tuning.
\newblock In Lun{-}Wei Ku, Andre Martins, and Vivek Srikumar, editors, \emph{Proceedings of the 62nd Annual Meeting of the Association for Computational Linguistics (Volume 1: Long Papers), {ACL} 2024, Bangkok, Thailand, August 11-16, 2024}, pages 11521--11567. Association for Computational Linguistics, 2024.
\newblock \doi{10.18653/V1/2024.ACL-LONG.620}.
\newblock URL \url{https://doi.org/10.18653/v1/2024.acl-long.620}.

\bibitem[Sturua et~al.(2024)Sturua, Mohr, Akram, G{\"{u}}nther, Wang, Krimmel, Wang, Mastrapas, Koukounas, Wang, and Xiao]{DBLP:journals/corr/abs-2409-10173}
Saba Sturua, Isabelle Mohr, Mohammad~Kalim Akram, Michael G{\"{u}}nther, Bo~Wang, Markus Krimmel, Feng Wang, Georgios Mastrapas, Andreas Koukounas, Nan Wang, and Han Xiao.
\newblock jina-embeddings-v3: Multilingual embeddings with task lora.
\newblock \emph{CoRR}, abs/2409.10173, 2024.
\newblock \doi{10.48550/ARXIV.2409.10173}.
\newblock URL \url{https://doi.org/10.48550/arXiv.2409.10173}.

\bibitem[Su et~al.(2023)Su, Shi, Kasai, Wang, Hu, Ostendorf, Yih, Smith, Zettlemoyer, and Yu]{DBLP:conf/acl/SuSKWHOYSZ023}
Hongjin Su, Weijia Shi, Jungo Kasai, Yizhong Wang, Yushi Hu, Mari Ostendorf, Wen{-}tau Yih, Noah~A. Smith, Luke Zettlemoyer, and Tao Yu.
\newblock One embedder, any task: Instruction-finetuned text embeddings.
\newblock In Anna Rogers, Jordan~L. Boyd{-}Graber, and Naoaki Okazaki, editors, \emph{Findings of the Association for Computational Linguistics: {ACL} 2023, Toronto, Canada, July 9-14, 2023}, pages 1102--1121. Association for Computational Linguistics, 2023.
\newblock \doi{10.18653/V1/2023.FINDINGS-ACL.71}.
\newblock URL \url{https://doi.org/10.18653/v1/2023.findings-acl.71}.

\bibitem[Su et~al.(2024)Su, Ahmed, Lu, Pan, Bo, and Liu]{DBLP:journals/ijon/SuALPBL24}
Jianlin Su, Murtadha H.~M. Ahmed, Yu~Lu, Shengfeng Pan, Wen Bo, and Yunfeng Liu.
\newblock Roformer: Enhanced transformer with rotary position embedding.
\newblock \emph{Neurocomputing}, 568:\penalty0 127063, 2024.
\newblock \doi{10.1016/J.NEUCOM.2023.127063}.
\newblock URL \url{https://doi.org/10.1016/j.neucom.2023.127063}.

\bibitem[Sun et~al.(2016)Sun, Li, Guo, Zhao, Zheng, Si, and Liu]{thuctc}
Maosong Sun, Jingyang Li, Zhipeng Guo, Yu~Zhao, Yabin Zheng, Xiance Si, and Zhiyuan Liu.
\newblock Thuctc: An efficient chinese text classifier, 2016.
\newblock URL \url{http://thuctc.thunlp.org/}.
\newblock Accessed: 2024-05-01.

\bibitem[Tan et~al.(2024)Tan, Li, Wang, Beigi, Jiang, Bhattacharjee, Karami, Li, Cheng, and Liu]{DBLP:conf/emnlp/TanLWBJBKL0024}
Zhen Tan, Dawei Li, Song Wang, Alimohammad Beigi, Bohan Jiang, Amrita Bhattacharjee, Mansooreh Karami, Jundong Li, Lu~Cheng, and Huan Liu.
\newblock Large language models for data annotation and synthesis: {A} survey.
\newblock In Yaser Al{-}Onaizan, Mohit Bansal, and Yun{-}Nung Chen, editors, \emph{Proceedings of the 2024 Conference on Empirical Methods in Natural Language Processing, {EMNLP} 2024, Miami, FL, USA, November 12-16, 2024}, pages 930--957. Association for Computational Linguistics, 2024.
\newblock URL \url{https://aclanthology.org/2024.emnlp-main.54}.

\bibitem[Tang et~al.(2021)Tang, Li, Liu, Hong, Wu, and Wang]{DBLP:conf/acl/TangL0H0020}
Hongxuan Tang, Hongyu Li, Jing Liu, Yu~Hong, Hua Wu, and Haifeng Wang.
\newblock Dureader{\_}robust: {A} chinese dataset towards evaluating robustness and generalization of machine reading comprehension in real-world applications.
\newblock In Chengqing Zong, Fei Xia, Wenjie Li, and Roberto Navigli, editors, \emph{Proceedings of the 59th Annual Meeting of the Association for Computational Linguistics and the 11th International Joint Conference on Natural Language Processing, {ACL/IJCNLP} 2021, (Volume 2: Short Papers), Virtual Event, August 1-6, 2021}, pages 955--963. Association for Computational Linguistics, 2021.
\newblock \doi{10.18653/V1/2021.ACL-SHORT.120}.
\newblock URL \url{https://doi.org/10.18653/v1/2021.acl-short.120}.

\bibitem[Tang et~al.(2016)Tang, Bai, and Ma]{ChineseSTS}
Shancheng Tang, Yunyue Bai, and Fuyu Ma.
\newblock Chinese semantic text similarity trainning dataset, 2016.
\newblock URL \url{https://github.com/IAdmireu/ChineseSTS}.
\newblock Accessed: 2024-05-01.

\bibitem[Thakur et~al.(2021)Thakur, Reimers, R{\"{u}}ckl{\'{e}}, Srivastava, and Gurevych]{DBLP:conf/nips/Thakur0RSG21}
Nandan Thakur, Nils Reimers, Andreas R{\"{u}}ckl{\'{e}}, Abhishek Srivastava, and Iryna Gurevych.
\newblock {BEIR:} {A} heterogeneous benchmark for zero-shot evaluation of information retrieval models.
\newblock In Joaquin Vanschoren and Sai{-}Kit Yeung, editors, \emph{Proceedings of the Neural Information Processing Systems Track on Datasets and Benchmarks 1, NeurIPS Datasets and Benchmarks 2021, December 2021, virtual}, 2021.
\newblock URL \url{https://datasets-benchmarks-proceedings.neurips.cc/paper/2021/hash/65b9eea6e1cc6bb9f0cd2a47751a186f-Abstract-round2.html}.

\bibitem[Thakur et~al.(2024)Thakur, Ni, {\'{A}}brego, Wieting, Lin, and Cer]{DBLP:conf/naacl/ThakurNAWLC24}
Nandan Thakur, Jianmo Ni, Gustavo~Hern{\'{a}}ndez {\'{A}}brego, John Wieting, Jimmy Lin, and Daniel Cer.
\newblock Leveraging llms for synthesizing training data across many languages in multilingual dense retrieval.
\newblock In Kevin Duh, Helena G{\'{o}}mez{-}Adorno, and Steven Bethard, editors, \emph{Proceedings of the 2024 Conference of the North American Chapter of the Association for Computational Linguistics: Human Language Technologies (Volume 1: Long Papers), {NAACL} 2024, Mexico City, Mexico, June 16-21, 2024}, pages 7699--7724. Association for Computational Linguistics, 2024.
\newblock \doi{10.18653/V1/2024.NAACL-LONG.426}.
\newblock URL \url{https://doi.org/10.18653/v1/2024.naacl-long.426}.

\bibitem[Thorne et~al.(2018)Thorne, Vlachos, Christodoulopoulos, and Mittal]{DBLP:conf/naacl/ThorneVCM18}
James Thorne, Andreas Vlachos, Christos Christodoulopoulos, and Arpit Mittal.
\newblock {FEVER:} a large-scale dataset for fact extraction and verification.
\newblock In Marilyn~A. Walker, Heng Ji, and Amanda Stent, editors, \emph{Proceedings of the 2018 Conference of the North American Chapter of the Association for Computational Linguistics: Human Language Technologies, {NAACL-HLT} 2018, New Orleans, Louisiana, USA, June 1-6, 2018, Volume 1 (Long Papers)}, pages 809--819. Association for Computational Linguistics, 2018.
\newblock \doi{10.18653/V1/N18-1074}.
\newblock URL \url{https://doi.org/10.18653/v1/n18-1074}.

\bibitem[Ustinian(2020)]{falv5983}
Ustinian.
\newblock Law question-answering dataset.
\newblock https://www.heywhale.com/mw/dataset/5e953ca8e7ec38002d02fca7, 2020.

\bibitem[Voorhees et~al.(2020)Voorhees, Alam, Bedrick, Demner{-}Fushman, Hersh, Lo, Roberts, Soboroff, and Wang]{DBLP:journals/sigir/VoorheesABDHLRS20}
Ellen~M. Voorhees, Tasmeer Alam, Steven Bedrick, Dina Demner{-}Fushman, William~R. Hersh, Kyle Lo, Kirk Roberts, Ian Soboroff, and Lucy~Lu Wang.
\newblock {TREC-COVID:} constructing a pandemic information retrieval test collection.
\newblock \emph{{SIGIR} Forum}, 54\penalty0 (1):\penalty0 1:1--1:12, 2020.
\newblock \doi{10.1145/3451964.3451965}.
\newblock URL \url{https://doi.org/10.1145/3451964.3451965}.

\bibitem[Wadden et~al.(2020)Wadden, Lin, Lo, Wang, van Zuylen, Cohan, and Hajishirzi]{DBLP:conf/emnlp/WaddenLLWZCH20}
David Wadden, Shanchuan Lin, Kyle Lo, Lucy~Lu Wang, Madeleine van Zuylen, Arman Cohan, and Hannaneh Hajishirzi.
\newblock Fact or fiction: Verifying scientific claims.
\newblock In Bonnie Webber, Trevor Cohn, Yulan He, and Yang Liu, editors, \emph{Proceedings of the 2020 Conference on Empirical Methods in Natural Language Processing, {EMNLP} 2020, Online, November 16-20, 2020}, pages 7534--7550. Association for Computational Linguistics, 2020.
\newblock \doi{10.18653/V1/2020.EMNLP-MAIN.609}.
\newblock URL \url{https://doi.org/10.18653/v1/2020.emnlp-main.609}.

\bibitem[Wang et~al.(2022)Wang, Yang, Huang, Jiao, Yang, Jiang, Majumder, and Wei]{DBLP:journals/corr/abs-2212-03533}
Liang Wang, Nan Yang, Xiaolong Huang, Binxing Jiao, Linjun Yang, Daxin Jiang, Rangan Majumder, and Furu Wei.
\newblock Text embeddings by weakly-supervised contrastive pre-training.
\newblock \emph{CoRR}, abs/2212.03533, 2022.
\newblock \doi{10.48550/ARXIV.2212.03533}.
\newblock URL \url{https://doi.org/10.48550/arXiv.2212.03533}.

\bibitem[Wang et~al.(2024{\natexlab{a}})Wang, Yang, Huang, Yang, Majumder, and Wei]{DBLP:conf/acl/WangYHYMW24}
Liang Wang, Nan Yang, Xiaolong Huang, Linjun Yang, Rangan Majumder, and Furu Wei.
\newblock Improving text embeddings with large language models.
\newblock In Lun{-}Wei Ku, Andre Martins, and Vivek Srikumar, editors, \emph{Proceedings of the 62nd Annual Meeting of the Association for Computational Linguistics (Volume 1: Long Papers), {ACL} 2024, Bangkok, Thailand, August 11-16, 2024}, pages 11897--11916. Association for Computational Linguistics, 2024{\natexlab{a}}.
\newblock \doi{10.18653/V1/2024.ACL-LONG.642}.
\newblock URL \url{https://doi.org/10.18653/v1/2024.acl-long.642}.

\bibitem[Wang et~al.(2024{\natexlab{b}})Wang, Yang, Huang, Yang, Majumder, and Wei]{DBLP:journals/corr/abs-2402-05672}
Liang Wang, Nan Yang, Xiaolong Huang, Linjun Yang, Rangan Majumder, and Furu Wei.
\newblock Multilingual {E5} text embeddings: {A} technical report.
\newblock \emph{CoRR}, abs/2402.05672, 2024{\natexlab{b}}.
\newblock \doi{10.48550/ARXIV.2402.05672}.
\newblock URL \url{https://doi.org/10.48550/arXiv.2402.05672}.

\bibitem[Wang et~al.(2020)Wang, Lo, Chandrasekhar, Reas, Yang, Eide, Funk, Kinney, Liu, Merrill, Mooney, Murdick, Rishi, Sheehan, Shen, Stilson, Wade, Wang, Wilhelm, Xie, Raymond, Weld, Etzioni, and Kohlmeier]{DBLP:journals/corr/abs-2004-10706}
Lucy~Lu Wang, Kyle Lo, Yoganand Chandrasekhar, Russell Reas, Jiangjiang Yang, Darrin Eide, Kathryn Funk, Rodney Kinney, Ziyang Liu, William Merrill, Paul Mooney, Dewey~A. Murdick, Devvret Rishi, Jerry Sheehan, Zhihong Shen, Brandon Stilson, Alex~D. Wade, Kuansan Wang, Chris Wilhelm, Boya Xie, Douglas Raymond, Daniel~S. Weld, Oren Etzioni, and Sebastian Kohlmeier.
\newblock {CORD-19:} the covid-19 open research dataset.
\newblock \emph{CoRR}, abs/2004.10706, 2020.
\newblock URL \url{https://arxiv.org/abs/2004.10706}.

\bibitem[Williams et~al.(2018)Williams, Nangia, and Bowman]{DBLP:conf/naacl/WilliamsNB18}
Adina Williams, Nikita Nangia, and Samuel~R. Bowman.
\newblock A broad-coverage challenge corpus for sentence understanding through inference.
\newblock In Marilyn~A. Walker, Heng Ji, and Amanda Stent, editors, \emph{Proceedings of the 2018 Conference of the North American Chapter of the Association for Computational Linguistics: Human Language Technologies, {NAACL-HLT} 2018, New Orleans, Louisiana, USA, June 1-6, 2018, Volume 1 (Long Papers)}, pages 1112--1122. Association for Computational Linguistics, 2018.
\newblock \doi{10.18653/V1/N18-1101}.
\newblock URL \url{https://doi.org/10.18653/v1/n18-1101}.

\bibitem[Xiao et~al.(2019)Xiao, Zhong, Guo, Tu, Liu, Sun, Zhang, Han, Hu, Wang, and Xu]{DBLP:journals/corr/abs-1911-08962}
Chaojun Xiao, Haoxi Zhong, Zhipeng Guo, Cunchao Tu, Zhiyuan Liu, Maosong Sun, Tianyang Zhang, Xianpei Han, Zhen Hu, Heng Wang, and Jianfeng Xu.
\newblock {CAIL2019-SCM:} {A} dataset of similar case matching in legal domain.
\newblock \emph{CoRR}, abs/1911.08962, 2019.
\newblock URL \url{http://arxiv.org/abs/1911.08962}.

\bibitem[Xiao et~al.(2024{\natexlab{a}})Xiao, Liu, Zhang, Muennighoff, Lian, and Nie]{DBLP:conf/sigir/XiaoLZMLN24}
Shitao Xiao, Zheng Liu, Peitian Zhang, Niklas Muennighoff, Defu Lian, and Jian{-}Yun Nie.
\newblock C-pack: Packed resources for general chinese embeddings.
\newblock In Grace~Hui Yang, Hongning Wang, Sam Han, Claudia Hauff, Guido Zuccon, and Yi~Zhang, editors, \emph{Proceedings of the 47th International {ACM} {SIGIR} Conference on Research and Development in Information Retrieval, {SIGIR} 2024, Washington DC, USA, July 14-18, 2024}, pages 641--649. {ACM}, 2024{\natexlab{a}}.
\newblock \doi{10.1145/3626772.3657878}.
\newblock URL \url{https://doi.org/10.1145/3626772.3657878}.

\bibitem[Xiao et~al.(2024{\natexlab{b}})Xiao, Liu, Zhang, and Xing]{DBLP:conf/acl/XiaoLZX24}
Shitao Xiao, Zheng Liu, Peitian Zhang, and Xingrun Xing.
\newblock Lm-cocktail: Resilient tuning of language models via model merging.
\newblock In Lun{-}Wei Ku, Andre Martins, and Vivek Srikumar, editors, \emph{Findings of the Association for Computational Linguistics, {ACL} 2024, Bangkok, Thailand and virtual meeting, August 11-16, 2024}, pages 2474--2488. Association for Computational Linguistics, 2024{\natexlab{b}}.
\newblock \doi{10.18653/V1/2024.FINDINGS-ACL.145}.
\newblock URL \url{https://doi.org/10.18653/v1/2024.findings-acl.145}.

\bibitem[Xie et~al.(2023)Xie, Dong, Wang, Lv, Yao, Gan, Wu, Li, Li, Liu, and Ma]{DBLP:conf/sigir/XieDWLYG0LL0M23}
Xiaohui Xie, Qian Dong, Bingning Wang, Feiyang Lv, Ting Yao, Weinan Gan, Zhijing Wu, Xiangsheng Li, Haitao Li, Yiqun Liu, and Jin Ma.
\newblock T2ranking: {A} large-scale chinese benchmark for passage ranking.
\newblock In Hsin{-}Hsi Chen, Wei{-}Jou~(Edward) Duh, Hen{-}Hsen Huang, Makoto~P. Kato, Josiane Mothe, and Barbara Poblete, editors, \emph{Proceedings of the 46th International {ACM} {SIGIR} Conference on Research and Development in Information Retrieval, {SIGIR} 2023, Taipei, Taiwan, July 23-27, 2023}, pages 2681--2690. {ACM}, 2023.
\newblock \doi{10.1145/3539618.3591874}.
\newblock URL \url{https://doi.org/10.1145/3539618.3591874}.

\bibitem[Xu et~al.(2020)Xu, Hu, Zhang, Li, Cao, Li, Xu, Sun, Yu, Yu, Tian, Dong, Liu, Shi, Cui, Li, Zeng, Wang, Xie, Li, Patterson, Tian, Zhang, Zhou, Liu, Zhao, Zhao, Yue, Zhang, Yang, Richardson, and Lan]{DBLP:conf/coling/XuHZLCLXSYYTDLS20}
Liang Xu, Hai Hu, Xuanwei Zhang, Lu~Li, Chenjie Cao, Yudong Li, Yechen Xu, Kai Sun, Dian Yu, Cong Yu, Yin Tian, Qianqian Dong, Weitang Liu, Bo~Shi, Yiming Cui, Junyi Li, Jun Zeng, Rongzhao Wang, Weijian Xie, Yanting Li, Yina Patterson, Zuoyu Tian, Yiwen Zhang, He~Zhou, Shaoweihua Liu, Zhe Zhao, Qipeng Zhao, Cong Yue, Xinrui Zhang, Zhengliang Yang, Kyle Richardson, and Zhenzhong Lan.
\newblock {CLUE:} {A} chinese language understanding evaluation benchmark.
\newblock In Donia Scott, N{\'{u}}ria Bel, and Chengqing Zong, editors, \emph{Proceedings of the 28th International Conference on Computational Linguistics, {COLING} 2020, Barcelona, Spain (Online), December 8-13, 2020}, pages 4762--4772. International Committee on Computational Linguistics, 2020.
\newblock \doi{10.18653/V1/2020.COLING-MAIN.419}.
\newblock URL \url{https://doi.org/10.18653/v1/2020.coling-main.419}.

\bibitem[Yadav et~al.(2023)Yadav, Tam, Choshen, Raffel, and Bansal]{DBLP:conf/nips/YadavTCRB23}
Prateek Yadav, Derek Tam, Leshem Choshen, Colin~A. Raffel, and Mohit Bansal.
\newblock Ties-merging: Resolving interference when merging models.
\newblock In Alice Oh, Tristan Naumann, Amir Globerson, Kate Saenko, Moritz Hardt, and Sergey Levine, editors, \emph{Advances in Neural Information Processing Systems 36: Annual Conference on Neural Information Processing Systems 2023, NeurIPS 2023, New Orleans, LA, USA, December 10 - 16, 2023}, 2023.
\newblock URL \url{http://papers.nips.cc/paper\_files/paper/2023/hash/1644c9af28ab7916874f6fd6228a9bcf-Abstract-Conference.html}.

\bibitem[Yang et~al.(2024)Yang, Yang, Hui, Zheng, Yu, Zhou, Li, Li, Liu, Huang, Dong, Wei, Lin, Tang, Wang, Yang, Tu, Zhang, Ma, Yang, Xu, Zhou, Bai, He, Lin, Dang, Lu, Chen, Yang, Li, Xue, Ni, Zhang, Wang, Peng, Men, Gao, Lin, Wang, Bai, Tan, Zhu, Li, Liu, Ge, Deng, Zhou, Ren, Zhang, Wei, Ren, Liu, Fan, Yao, Zhang, Wan, Chu, Liu, Cui, Zhang, Guo, and Fan]{DBLP:journals/corr/abs-2407-10671}
An~Yang, Baosong Yang, Binyuan Hui, Bo~Zheng, Bowen Yu, Chang Zhou, Chengpeng Li, Chengyuan Li, Dayiheng Liu, Fei Huang, Guanting Dong, Haoran Wei, Huan Lin, Jialong Tang, Jialin Wang, Jian Yang, Jianhong Tu, Jianwei Zhang, Jianxin Ma, Jianxin Yang, Jin Xu, Jingren Zhou, Jinze Bai, Jinzheng He, Junyang Lin, Kai Dang, Keming Lu, Keqin Chen, Kexin Yang, Mei Li, Mingfeng Xue, Na~Ni, Pei Zhang, Peng Wang, Ru~Peng, Rui Men, Ruize Gao, Runji Lin, Shijie Wang, Shuai Bai, Sinan Tan, Tianhang Zhu, Tianhao Li, Tianyu Liu, Wenbin Ge, Xiaodong Deng, Xiaohuan Zhou, Xingzhang Ren, Xinyu Zhang, Xipin Wei, Xuancheng Ren, Xuejing Liu, Yang Fan, Yang Yao, Yichang Zhang, Yu~Wan, Yunfei Chu, Yuqiong Liu, Zeyu Cui, Zhenru Zhang, Zhifang Guo, and Zhihao Fan.
\newblock Qwen2 technical report.
\newblock \emph{CoRR}, abs/2407.10671, 2024.
\newblock \doi{10.48550/ARXIV.2407.10671}.
\newblock URL \url{https://doi.org/10.48550/arXiv.2407.10671}.

\bibitem[Yang et~al.(2023)Yang, Yuan, Fan, Yang, Wang, Wang, and Zhao]{DBLP:conf/emnlp/YangYFYWWZ23}
Dongjie Yang, Ruifeng Yuan, Yuantao Fan, Yifei Yang, Zili Wang, Shusen Wang, and Hai Zhao.
\newblock Refgpt: Dialogue generation of gpt, by gpt, and for {GPT}.
\newblock In Houda Bouamor, Juan Pino, and Kalika Bali, editors, \emph{Findings of the Association for Computational Linguistics: {EMNLP} 2023, Singapore, December 6-10, 2023}, pages 2511--2535. Association for Computational Linguistics, 2023.
\newblock \doi{10.18653/V1/2023.FINDINGS-EMNLP.165}.
\newblock URL \url{https://doi.org/10.18653/v1/2023.findings-emnlp.165}.

\bibitem[Yang et~al.(2019)Yang, Zhang, Tar, and Baldridge]{DBLP:conf/emnlp/YangZTB19}
Yinfei Yang, Yuan Zhang, Chris Tar, and Jason Baldridge.
\newblock {PAWS-X:} {A} cross-lingual adversarial dataset for paraphrase identification.
\newblock In Kentaro Inui, Jing Jiang, Vincent Ng, and Xiaojun Wan, editors, \emph{Proceedings of the 2019 Conference on Empirical Methods in Natural Language Processing and the 9th International Joint Conference on Natural Language Processing, {EMNLP-IJCNLP} 2019, Hong Kong, China, November 3-7, 2019}, pages 3685--3690. Association for Computational Linguistics, 2019.
\newblock \doi{10.18653/V1/D19-1382}.
\newblock URL \url{https://doi.org/10.18653/v1/D19-1382}.

\bibitem[Yang et~al.(2018)Yang, Qi, Zhang, Bengio, Cohen, Salakhutdinov, and Manning]{DBLP:conf/emnlp/Yang0ZBCSM18}
Zhilin Yang, Peng Qi, Saizheng Zhang, Yoshua Bengio, William~W. Cohen, Ruslan Salakhutdinov, and Christopher~D. Manning.
\newblock Hotpotqa: {A} dataset for diverse, explainable multi-hop question answering.
\newblock In Ellen Riloff, David Chiang, Julia Hockenmaier, and Jun'ichi Tsujii, editors, \emph{Proceedings of the 2018 Conference on Empirical Methods in Natural Language Processing, Brussels, Belgium, October 31 - November 4, 2018}, pages 2369--2380. Association for Computational Linguistics, 2018.
\newblock \doi{10.18653/V1/D18-1259}.
\newblock URL \url{https://doi.org/10.18653/v1/d18-1259}.

\bibitem[Yu et~al.(2024)Yu, Yu, Yu, Huang, and Li]{DBLP:conf/icml/Yu0Y0L24}
Le~Yu, Bowen Yu, Haiyang Yu, Fei Huang, and Yongbin Li.
\newblock Language models are super mario: Absorbing abilities from homologous models as a free lunch.
\newblock In \emph{Forty-first International Conference on Machine Learning, {ICML} 2024, Vienna, Austria, July 21-27, 2024}. OpenReview.net, 2024.
\newblock URL \url{https://openreview.net/forum?id=fq0NaiU8Ex}.

\bibitem[Yuan et~al.(2021)Yuan, Zhao, Du, Ding, Liu, Cen, Zou, Yang, and Tang]{DBLP:journals/aiopen/YuanZDDLCZYT21}
Sha Yuan, Hanyu Zhao, Zhengxiao Du, Ming Ding, Xiao Liu, Yukuo Cen, Xu~Zou, Zhilin Yang, and Jie Tang.
\newblock Wudaocorpora: {A} super large-scale chinese corpora for pre-training language models.
\newblock \emph{{AI} Open}, 2:\penalty0 65--68, 2021.
\newblock \doi{10.1016/J.AIOPEN.2021.06.001}.
\newblock URL \url{https://doi.org/10.1016/j.aiopen.2021.06.001}.

\bibitem[Zhang et~al.(2018)Zhang, Zhang, Wang, Guo, and Liu]{DBLP:journals/access/ZhangZWGL18}
Sheng Zhang, Xin Zhang, Hui Wang, Lixiang Guo, and Shanshan Liu.
\newblock Multi-scale attentive interaction networks for chinese medical question answer selection.
\newblock \emph{{IEEE} Access}, 6:\penalty0 74061--74071, 2018.
\newblock \doi{10.1109/ACCESS.2018.2883637}.
\newblock URL \url{https://doi.org/10.1109/ACCESS.2018.2883637}.

\bibitem[Zhang et~al.(2022)Zhang, Liang, Gong, Jiang, and Duan]{DBLP:conf/acl/ZhangLGJD22}
Shunyu Zhang, Yaobo Liang, Ming Gong, Daxin Jiang, and Nan Duan.
\newblock Multi-view document representation learning for open-domain dense retrieval.
\newblock In Smaranda Muresan, Preslav Nakov, and Aline Villavicencio, editors, \emph{Proceedings of the 60th Annual Meeting of the Association for Computational Linguistics (Volume 1: Long Papers), {ACL} 2022, Dublin, Ireland, May 22-27, 2022}, pages 5990--6000. Association for Computational Linguistics, 2022.
\newblock \doi{10.18653/V1/2022.ACL-LONG.414}.
\newblock URL \url{https://doi.org/10.18653/v1/2022.acl-long.414}.

\bibitem[Zhang et~al.(2024)Zhang, Zhang, Long, Xie, Dai, Tang, Lin, Yang, Xie, Huang, Zhang, Li, and Zhang]{DBLP:conf/emnlp/ZhangZLXDTLYXHZ24}
Xin Zhang, Yanzhao Zhang, Dingkun Long, Wen Xie, Ziqi Dai, Jialong Tang, Huan Lin, Baosong Yang, Pengjun Xie, Fei Huang, Meishan Zhang, Wenjie Li, and Min Zhang.
\newblock mgte: Generalized long-context text representation and reranking models for multilingual text retrieval.
\newblock In Franck Dernoncourt, Daniel Preotiuc{-}Pietro, and Anastasia Shimorina, editors, \emph{Proceedings of the 2024 Conference on Empirical Methods in Natural Language Processing: {EMNLP} 2024 - Industry Track, Miami, Florida, USA, November 12-16, 2024}, pages 1393--1412. Association for Computational Linguistics, 2024.
\newblock URL \url{https://aclanthology.org/2024.emnlp-industry.103}.

\bibitem[Zhang et~al.(2021)Zhang, Ma, Shi, and Lin]{DBLP:journals/corr/abs-2108-08787}
Xinyu Zhang, Xueguang Ma, Peng Shi, and Jimmy Lin.
\newblock Mr. tydi: {A} multi-lingual benchmark for dense retrieval.
\newblock \emph{CoRR}, abs/2108.08787, 2021.
\newblock URL \url{https://arxiv.org/abs/2108.08787}.

\bibitem[Zhang et~al.(2023)Zhang, Thakur, Ogundepo, Kamalloo, Alfonso{-}Hermelo, Li, Liu, Rezagholizadeh, and Lin]{DBLP:journals/tacl/0018TOKAL0RL23}
Xinyu Zhang, Nandan Thakur, Odunayo Ogundepo, Ehsan Kamalloo, David Alfonso{-}Hermelo, Xiaoguang Li, Qun Liu, Mehdi Rezagholizadeh, and Jimmy Lin.
\newblock {MIRACL:} {A} multilingual retrieval dataset covering 18 diverse languages.
\newblock \emph{Trans. Assoc. Comput. Linguistics}, 11:\penalty0 1114--1131, 2023.
\newblock \doi{10.1162/TACL\_A\_00595}.
\newblock URL \url{https://doi.org/10.1162/tacl\_a\_00595}.

\bibitem[Zhao et~al.(2024)Zhao, Zhong, Sun, Hu, Liu, Li, Hu, and Zhang]{DBLP:journals/corr/abs-2410-10293}
Xinping Zhao, Yan Zhong, Zetian Sun, Xinshuo Hu, Zhenyu Liu, Dongfang Li, Baotian Hu, and Min Zhang.
\newblock Funnelrag: {A} coarse-to-fine progressive retrieval paradigm for {RAG}.
\newblock \emph{CoRR}, abs/2410.10293, 2024.
\newblock \doi{10.48550/ARXIV.2410.10293}.
\newblock URL \url{https://doi.org/10.48550/arXiv.2410.10293}.

\bibitem[Zhao et~al.(2022)Zhao, Liu, Zhang, Tang, and Hu]{DBLP:journals/corr/abs-2202-10974}
Zhen Zhao, Yuqiu Liu, Gang Zhang, Liang Tang, and Xiaolin Hu.
\newblock The winning solution to the iflytek challenge 2021 cultivated land extraction from high-resolution remote sensing image.
\newblock \emph{CoRR}, abs/2202.10974, 2022.
\newblock URL \url{https://arxiv.org/abs/2202.10974}.

\bibitem[Zheng et~al.(2024)Zheng, Zhang, Shen, Liu, Lin, Fu, Chen, and Yue]{DBLP:conf/acl/ZhengZSLLFCY24}
Tianyu Zheng, Ge~Zhang, Tianhao Shen, Xueling Liu, Bill~Yuchen Lin, Jie Fu, Wenhu Chen, and Xiang Yue.
\newblock Opencodeinterpreter: Integrating code generation with execution and refinement.
\newblock In Lun{-}Wei Ku, Andre Martins, and Vivek Srikumar, editors, \emph{Findings of the Association for Computational Linguistics, {ACL} 2024, Bangkok, Thailand and virtual meeting, August 11-16, 2024}, pages 12834--12859. Association for Computational Linguistics, 2024.
\newblock \doi{10.18653/V1/2024.FINDINGS-ACL.762}.
\newblock URL \url{https://doi.org/10.18653/v1/2024.findings-acl.762}.

\bibitem[Zhou et~al.(2023)Zhou, Liu, Xu, Iyer, Sun, Mao, Ma, Efrat, Yu, Yu, Zhang, Ghosh, Lewis, Zettlemoyer, and Levy]{DBLP:conf/nips/ZhouLX0SMMEYYZG23}
Chunting Zhou, Pengfei Liu, Puxin Xu, Srinivasan Iyer, Jiao Sun, Yuning Mao, Xuezhe Ma, Avia Efrat, Ping Yu, Lili Yu, Susan Zhang, Gargi Ghosh, Mike Lewis, Luke Zettlemoyer, and Omer Levy.
\newblock {LIMA:} less is more for alignment.
\newblock In Alice Oh, Tristan Naumann, Amir Globerson, Kate Saenko, Moritz Hardt, and Sergey Levine, editors, \emph{Advances in Neural Information Processing Systems 36: Annual Conference on Neural Information Processing Systems 2023, NeurIPS 2023, New Orleans, LA, USA, December 10 - 16, 2023}, 2023.
\newblock URL \url{http://papers.nips.cc/paper\_files/paper/2023/hash/ac662d74829e4407ce1d126477f4a03a-Abstract-Conference.html}.

\bibitem[Zhou et~al.(2024)Zhou, Dai, Cao, Zhang, and Xu]{DBLP:journals/corr/abs-2410-24200}
Yuqi Zhou, Sunhao Dai, Zhanshuo Cao, Xiao Zhang, and Jun Xu.
\newblock Length-induced embedding collapse in transformer-based models.
\newblock \emph{CoRR}, abs/2410.24200, 2024.
\newblock \doi{10.48550/ARXIV.2410.24200}.
\newblock URL \url{https://doi.org/10.48550/arXiv.2410.24200}.

\bibitem[Zhu(2023)]{ChatMed}
Wei Zhu.
\newblock Chatmed-dataset: An gpt generated medical query-response datasets for medcial large language models.
\newblock https://github.com/michael-wzhu/ChatMed, 2023.

\end{thebibliography}

\appendix

\section{Appendix}
We mainly compare the multilingual models paraphrase-multilingual-mpnet-base-v2~\citep{DBLP:conf/emnlp/ReimersG19}, multilingual-e5-large~\citep{DBLP:journals/corr/abs-2402-05672}, bge-m3~\citep{DBLP:journals/corr/abs-2402-03216}, and gte-multilingual-base~\citep{DBLP:conf/emnlp/ZhangZLXDTLYXHZ24}, along with some models that demonstrated multilingual results such as Cohere-embed-multilingual-v3.0, jina-embeddings-v3~\citep{DBLP:journals/corr/abs-2409-10173}, and e5-mistral-7b-instruct~\citep{DBLP:conf/acl/WangYHYMW24}. The detailed results on MTEB of different language is presented in \autoref{tab:mteb_detail_zh}, \autoref{tab:mteb_detail_en}, \autoref{tab:mteb_detail_fr} and \autoref{tab:mteb_detail_pl}.

\begin{table}[H]
  \centering
  \scriptsize
  \begin{tabular}{l|ccccccc}
    \toprule
    \multirow{2}{*}{\textbf{Model}} &   \multicolumn{7}{c}{\textbf{MTEB (zh)}}  \\
    \cmidrule(r){2-8}
               & \textbf{avg}  & \textbf{Class.} & \textbf{Clust.} & \textbf{PairCl.} & \textbf{Reran.} & \textbf{Retri.} & \textbf{STS} \\
    \hline
    
    e5-mistral-7b-instruct         & 60.89 & 70.47 & 52.30 & 72.19 & 61.86 & 61.75 & 50.22 \\ 

    paraphrase-multilingual-mpnet-base-v2 & 44.59 & 62.7 & 39.67 & 80.90 & 44.91 & 22.92 & 39.11 \\
    multilingual-e5-large             & 58.54 & 66.29 & 48.23 & 69.89 & 56.00 & 63.66 & 48.29 \\
    bge-m3 (Dense)                   & 60.80 & 66.95 & 45.75 & 73.98 & 62.88 & 65.43 & 52.43 \\
    gte-multilingual-base (Dense)    & 62.72 & 64.27 & 47.48 & 78.34 & 68.17 & 71.95 & 52.73 \\
    \rowcolor{mygray}
    \textbf{KaLM-embedding-mini-instruct}   & 64.13 & 70.94 & 57.32 & 72.94 & 64.38 & 70.11 & 51.57 \\
    \bottomrule
  \end{tabular}
  \caption{Embedding model performance on MTEB Chinese (C-MTEB).}
  \label{tab:mteb_detail_zh}
\end{table}

\begin{table}[H]
  \centering
  \scriptsize
  \begin{tabular}{l|cccccccc}
    \toprule
    \multirow{2}{*}{\textbf{Model}} &   \multicolumn{8}{c}{\textbf{MTEB (en)}}  \\
    \cmidrule(r){2-9}
               & \textbf{avg}  & \textbf{Class.} & \textbf{Clust.} & \textbf{PairCl.} & \textbf{Reran.} & \textbf{Retri.} & \textbf{STS} & \textbf{Summ.} \\
    \hline
    Cohere-embed-multilingual-v3.0 & 64.01 & 76.01 & 46.60 & 86.15 & 57.86 & 53.84 & 83.15 & 30.99 \\
    jina-embeddings-v3 (Multi-LoRA) & 65.51 & 82.58 & 45.21 & 84.01 & 58.13 & 53.88 & 85.81 & 29.71 \\
    e5-mistral-7b-instruct          & 66.63 & 78.47 & 50.26 & 88.34 & 60.21 & 56.89 & 84.63 & 31.40 \\ 

    paraphrase-multilingual-mpnet-base-v2 & 54.64 & 67.46 & 38.50 & 80.81 & 53.80 & 35.34 & 80.77 & 31.57 \\
    multilingual-e5-large             & 60.89 & 71.77 & 41.23 & 84.75 & 55.96 & 51.40 & 81.62 & 29.64 \\
    bge-m3 (Dense)                    & 59.84 & 74.08 & 37.27 & 84.50 & 55.28 & 48.82 & 81.37 & 31.55 \\
    gte-multilingual-base (Dense)    & 61.40 & 70.89 & 44.31 & 84.23 & 57.47 & 51.08 & 82.11 & 30.58 \\
    \rowcolor{mygray}
    \textbf{KaLM-embedding-mini-instruct}   & 64.94 & 84.74 & 47.82 & 83.26 & 55.41 & 51.65 & 82.24 & 25.23 \\
    \bottomrule
  \end{tabular}
  \caption{Embedding model performance on MTEB English.}
  \label{tab:mteb_detail_en}
\end{table}

\begin{table}[H]
  \centering
  \scriptsize
  \begin{tabular}{l|cccccccc}
    \toprule
    \multirow{2}{*}{\textbf{Model}} &   \multicolumn{8}{c}{\textbf{MTEB (fr)}}  \\
    \cmidrule(r){2-9}
               & \textbf{avg}  & \textbf{Class.} & \textbf{Clust.} & \textbf{PairCl.} & \textbf{Reran.} & \textbf{Retri.} & \textbf{STS} & \textbf{Summ.} \\
    \hline
    Cohere-embed-multilingual-v3.0 & 56.02 & 67.08 & 40.70 & 77.67 & 68.36 & 40.42 & 81.28 & 31.26 \\
    jina-embeddings-v3 (Multi-LoRA) & 62.29 & 76.54 & 44.95 & 76.60 & 69.85 & 53.48 & 84.89 & 30.71 \\
    e5-mistral-7b-instruct          & 48.33 & 57.72 & 41.16 & 76.08 & 62.2 & 23.44 & 65.36 & 32.22 \\ 

    paraphrase-multilingual-mpnet-base-v2 & 55.21 & 64.73 & 41.79 & 75.80 & 74.09 & 38.13 & 78.18 & 29.47 \\
    multilingual-e5-large             & 55.64 & 66.54 & 38.70 & 76.19 & 72.14 & 42.17 & 79.36 & 30.92 \\
    bge-m3 (Dense)                    & 58.79 & 71.57 & 36.54 & 79.78 & 77.36 & 51.13 & 80.78 & 31.05 \\
    gte-multilingual-base (Dense)    & 59.79 & 68.72 & 41.66 & 79.47 & 76.47 & 52.97 & 81.36 & 29.74 \\
    \rowcolor{mygray}
    \textbf{KaLM-embedding-mini-instruct}   & 63.08 & 75.46 & 51.92 & 76.88 & 80.58 & 48.37 & 79.25 & 29.32 \\
    \bottomrule
  \end{tabular}
  \caption{Embedding model performance on MTEB French.}
  \label{tab:mteb_detail_fr}
\end{table}

\begin{table}[H]
  \centering
  \scriptsize
  \begin{tabular}{l|cccccc}
    \toprule
    \multirow{2}{*}{\textbf{Model}} &   \multicolumn{6}{c}{\textbf{MTEB (pl)}}  \\
    \cmidrule(r){2-7}
               & \textbf{avg}  & \textbf{Class.} & \textbf{Clust.} & \textbf{PairCl.} & \textbf{Retri.} & \textbf{STS} \\
    \hline
    jina-embeddings-v3 (Multi-LoRA) & 63.97 & 70.81 & 43.66 & 83.70 & 51.89 & 72.77 \\

    paraphrase-multilingual-mpnet-base-v2 & 	48.67 & 54.08 & 25.62 & 86.23 & 29.17 & 65.19 \\
    multilingual-e5-large             & 60.08 & 63.82 & 33.88 & 85.5 & 48.98 & 66.91 \\
    bge-m3 (Dense)                    & 60.35 & 65.15 & 25.21 & 86.46 & 48.51 & 69.44 \\
    gte-multilingual-base (Dense)    & 58.22 & 60.15 & 33.67 & 85.45 & 46.40 & 68.92 \\
    \rowcolor{mygray}
    \textbf{KaLM-embedding-mini-instruct}   & 57.05 & 64.50 & 39.24 & 79.01 & 43.24 & 67.02 \\
    \bottomrule
  \end{tabular}
  \caption{Embedding model performance on MTEB Polish.}
  \label{tab:mteb_detail_pl}
\end{table}


\begin{figure}[htbp]
    \centering
    \begin{subfigure}[b]{0.495\textwidth}
        \centering
        \includegraphics[width=\textwidth]{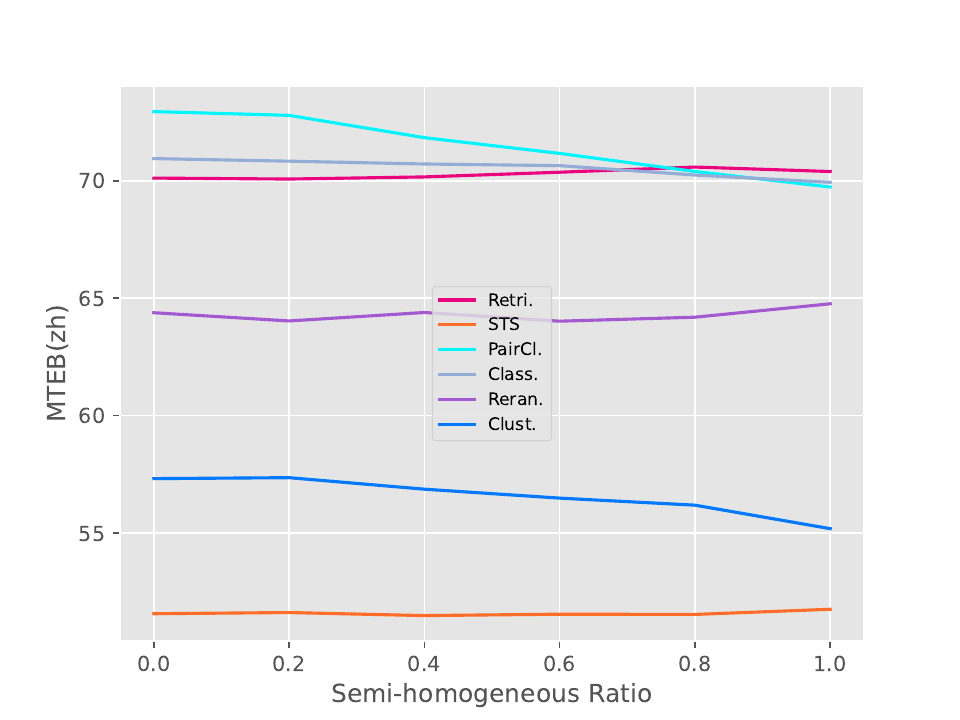}
        \subcaption{MTEB(zh)}
        \label{fig:semi-homogeneous_task_batch_trend_detailed_tasks_zh}
    \end{subfigure}
    \hfill
    \begin{subfigure}[b]{0.495\textwidth}
        \centering
        \includegraphics[width=\textwidth]{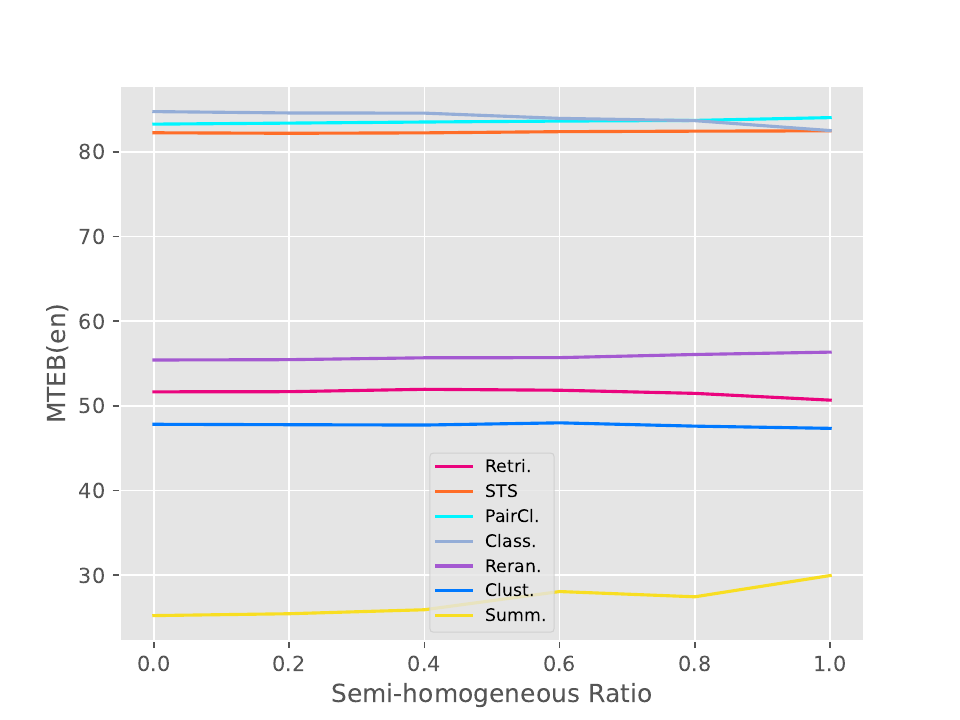}
        \subcaption{MTEB(en)}
        \label{fig:semi-homogeneous_task_batch_trend_detailed_tasks_en}
    \end{subfigure}
    \caption{Impact of semi-homogeneous task batch on detailed tasks from MTEB in Chinese and English.}
    \label{fig:semi-homogeneous_task_batch_trend_detailed_tasks}
\end{figure}

\begin{table}[htbp]
  \centering
  \small
  \begin{tabular}{lcc}
    \toprule
    Source              & Language      &  Pairs  \\
    \hline
    \href{https://huggingface.co/datasets/McAuley-Lab/Amazon-Reviews-2023}{Amazon-Reviews}~\citep{DBLP:journals/corr/abs-2403-03952}     
    & multilingual  
    & 23M \\
    
    \href{https://huggingface.co/datasets/intfloat/multilingual_cc_news}{CC-News}~\citep{DBLP:conf/isiwi/HamborgMBG17}      
    & multilingual  
    & 100M  \\
    
    \href{https://huggingface.co/datasets/allenai/nllb}{NLLB}~\citep{DBLP:journals/corr/abs-2207-04672,DBLP:conf/emnlp/HeffernanCS22,DBLP:conf/acl/SchwenkWEGJF20}          
    & multilingual  
    & 2M  \\
    
    \href{https://huggingface.co/datasets/Cohere/wikipedia-2023-11-embed-multilingual-v3}{Wikipedia}~\citep{wikidump}       
    & multilingual  
    & 100M  \\
    
    \href{https://huggingface.co/datasets/bigscience/xP3}{xP3}~\citep{DBLP:conf/acl/MuennighoffWSRB23}          
    & multilingual  
    & 19M  \\
    
    \href{https://huggingface.co/datasets/GEM/xlsum}{XL-Sum}~\citep{DBLP:conf/acl/HasanBIMLKRS21}
    & multilingual  
    & 1M  \\
    
    \href{https://huggingface.co/datasets/nthakur/swim-ir-monolingual}{SWIM-IR (Monolingual)}~\citep{DBLP:conf/naacl/ThakurNAWLC24} 
    & multilingual  
    & 3M  \\
    
    \href{https://huggingface.co/datasets/nthakur/swim-ir-cross-lingual}{SWIM-IR (Cross-lingual)}~\citep{DBLP:conf/naacl/ThakurNAWLC24}   
    & multilingual  
    & 15M \\
    
    \href{https://huggingface.co/datasets/neuclir/csl}{CSL}~\citep{DBLP:conf/coling/LiZ0S0MZ22}             
    & zh            
    & 0.4M \\
    
    \href{https://data.baai.ac.cn/details/WuDaoCorporaText}{Wudao}~\citep{DBLP:journals/aiopen/YuanZDDLCZYT21}
    & zh            
    & 44M \\
    
    \href{https://huggingface.co/datasets/SirlyDreamer/THUCNews}{THUCNews}~\citep{thuctc}
    & zh            
    & 0.8M  \\
    
    \href{https://huggingface.co/datasets/wangrui6/Zhihu-KOL}{Zhihu-KOL}         
    & zh            
    & 0.8M \\
    
    \href{https://huggingface.co/datasets/sentence-transformers/codesearchnet}{CodeSearchNet}~\citep{DBLP:journals/corr/abs-1909-09436}   
    & en           
    & 1M \\
    
    \href{https://huggingface.co/datasets/sentence-transformers/paq}{PAQ}~\citep{DBLP:journals/tacl/LewisWLMKPSR21}            
    & en            
    & 9M \\
    
    \href{https://huggingface.co/datasets/sentence-transformers/reddit}{Reddit}          
    & en            
    & 100M \\
    
    \href{https://huggingface.co/datasets/teven/stackexchange}{StackExchange}   
    & en            
    & 14M \\
    
    \href{https://huggingface.co/datasets/sentence-transformers/s2orc}{S2ORC}           
    & en            
    & 41M \\
    \bottomrule
  \end{tabular}
  \caption{Pre-training data list.}
  \label{tab:pretrain_data_list}
\end{table}

\newpage

\newpage
\thispagestyle{empty}

\begin{table}[htbp]
    \centering
    \tiny

    \begin{tabular}{l|ccccc}
        \toprule
        Source & Type & Categ. & Language      &  Pairs  & Pairs(filtered) \\
        \hline
        \hline
        
        \href{https://huggingface.co/datasets/m-a-p/CodeFeedback-Filtered-Instruction}{CodeFeedback}~\citep{DBLP:conf/acl/ZhengZSLLFCY24}   
        & Retrieval
        & s2p  
        & en
        & 50000 
        & 49090 \\

        \href{https://huggingface.co/datasets/rusano/ELI5_custom}{ELI5}~\citep{DBLP:conf/acl/FanJPGWA19}   
        & Retrieval
        & s2p  
        & en
        & 100000 
        & 76408 \\

        \href{https://github.com/chaitanyamalaviya/ExpertQA}{ExpertQA}~\citep{DBLP:conf/naacl/MalaviyaLCSYR24}   
        & Retrieval
        & s2p  
        & en
        & 1261 
        & 1252 \\

        \href{https://github.com/allenai/gooaq}{GooAQ}~\citep{DBLP:conf/emnlp/KhashabiNKSHC21}   
        & Retrieval
        & s2p  
        & en
        & 50000 
        & 49833 \\

        \href{https://hf.co/datasets/GritLM/MEDI2BGE}{MEDI2BGE}~\citep{DBLP:journals/corr/abs-2402-09906,DBLP:conf/acl/SuSKWHOYSZ023}   
        & Retrieval
        & s2p  
        & en
        & 100000 
        & 71790 \\

        \href{https://huggingface.co/datasets/Open-Orca/OpenOrca}{OpenOrca}~\citep{DBLP:journals/corr/abs-2306-02707}   
        & Retrieval
        & s2p  
        & en
        & 40000 
        & 38623 \\

        \href{https://huggingface.co/datasets/sentence-transformers/paq}{PAQ}~\citep{DBLP:journals/tacl/LewisWLMKPSR21} 
        & Retrieval
        & s2p  
        & en
        & 50000 
        & 49849 \\

        \href{https://huggingface.co/datasets/qiaojin/PubMedQA}{PubMedQA}~\citep{DBLP:conf/emnlp/JinDLCL19}   
        & Retrieval
        & s2p  
        & en
        & 80000 
        & 79954 \\

        \href{https://huggingface.co/datasets/kyunghyuncho/search_qa}{SearchQA}~\citep{DBLP:journals/corr/DunnSHGCC17}   
        & Retrieval
        & s2p  
        & en
        & 10000 
        & 9988 \\

        \href{https://huggingface.co/datasets/TitanMLData/arxiv_qa}{arxiv\_qa}   
        & Retrieval
        & s2p  
        & en
        & 23397 
        & 17927 \\

        \href{https://huggingface.co/datasets/intfloat/multilingual_cc_news}{CC-News}~\citep{DBLP:conf/isiwi/HamborgMBG17}
        & Retrieval
        & s2p  
        & en
        & 30000 
        & 28246 \\

        \href{https://huggingface.co/datasets/irds/cord19_trec-covid}{TREC-COVID}~\citep{DBLP:journals/sigir/VoorheesABDHLRS20,DBLP:journals/corr/abs-2004-10706}   
        & Retrieval
        & s2p  
        & en
        & 50000 
        & 48517 \\

        \href{https://huggingface.co/datasets/BeIR/dbpedia-entity-generated-queries}{DBpedia-Entity}~\citep{DBLP:conf/nips/Thakur0RSG21}   
        & Retrieval
        & s2p  
        & en
        & 100000 
        & 96792 \\

        \href{https://huggingface.co/datasets/tasksource/esci}{ESCI}~\citep{DBLP:journals/corr/abs-2206-06588}   
        & Retrieval
        & s2p  
        & en
        & 30000 
        & 26043 \\

        \href{https://huggingface.co/datasets/maxzoech/fever}{FEVER}~\citep{DBLP:conf/naacl/ThorneVCM18}   
        & Retrieval
        & s2p  
        & en
        & 87855 
        & 87216 \\

        \href{https://huggingface.co/datasets/irds/beir_fiqa_train}{FiQA}~\citep{DBLP:conf/www/MaiaHFDMZB18}   
        & Retrieval
        & s2p  
        & en
        & 5490 
        & 4689 \\

        \href{https://huggingface.co/datasets/hotpotqa/hotpot_qa}{HotpotQA}~\citep{DBLP:conf/emnlp/Yang0ZBCSM18}   
        & Retrieval
        & s2p  
        & en
        & 184057 
        & 150153 \\

        \href{https://huggingface.co/datasets/Shitao/MLDR}{MLDR}~\citep{DBLP:journals/corr/abs-2402-03216}   
        & Retrieval
        & s2p  
        & en
        & 41434 
        & 31097 \\

        \href{https://huggingface.co/datasets/Tevatron/msmarco-passage}{MSMARCO}~\citep{DBLP:conf/nips/NguyenRSGTMD16}    
        & Retrieval
        & s2p  
        & en
        & 175133 
        & 174190 \\

        \href{https://huggingface.co/datasets/mteb/msmarco-v2}{MSMARCO-v2}~\citep{DBLP:conf/nips/NguyenRSGTMD16}   
        & Retrieval
        & s2p  
        & en
        & 277144 
        & 258617 \\

        \href{https://huggingface.co/datasets/BeIR/nfcorpus-generated-queries}{NFCorpus}~\citep{DBLP:conf/ecir/BotevaGSR16}   
        & Retrieval
        & s2p  
        & en
        & 10824 
        & 10471 \\

        \href{https://huggingface.co/datasets/neural-bridge/rag-dataset-12000}{rag-dataset-12000}   
        & Retrieval
        & s2p  
        & en
        & 9590 
        & 9272 \\

        \href{https://huggingface.co/datasets/Tevatron/scifact}{SciFact}~\citep{DBLP:conf/emnlp/WaddenLLWZCH20}   
        & Retrieval
        & s2p  
        & en
        & 809 
        & 794 \\

        \href{https://huggingface.co/datasets/rajpurkar/squad_v2}{SQuAD 2.0}~\citep{DBLP:conf/acl/RajpurkarJL18,DBLP:conf/emnlp/RajpurkarZLL16}   
        & Retrieval
        & s2p  
        & en
        & 130217 
        & 125816 \\

        \href{https://huggingface.co/datasets/multi-train/emb-triviaqa-train}{TriviaQA}~\citep{DBLP:conf/acl/JoshiCWZ17}   
        & Retrieval
        & s2p  
        & en
        & 52886 
        & 44442 \\

        \href{https://huggingface.co/datasets/openai/webgpt_comparisons}{WebGPT Comparisons}~\citep{DBLP:journals/corr/abs-2112-09332}   
        & Retrieval
        & s2p  
        & en
        & 19242 
        & 18924 \\

        \href{https://huggingface.co/datasets/Tevatron/wikipedia-nq}{Natural Questions}~\citep{DBLP:journals/tacl/KwiatkowskiPRCP19}   
        & Retrieval
        & s2p  
        & en
        & 58622 
        & 56377 \\

        \href{https://huggingface.co/datasets/sentence-transformers/yahoo-answers}{Yahoo Answers}   
        & Retrieval
        & s2p  
        & en
        & 30000 
        & 21724 \\

        \href{https://huggingface.co/datasets/kiddothe2b/contract-nli}{ContractNLI}~\citep{DBLP:conf/emnlp/KoreedaM21}   
        & STS
        & s2s  
        & en
        & 3195 
        & 628 \\

        \href{https://huggingface.co/datasets/SetFit/mnli}{MultiNLI}~\citep{DBLP:conf/naacl/WilliamsNB18}   
        & STS
        & s2s  
        & en
        & 64674 
        & 63701 \\

        \href{https://huggingface.co/datasets/breakend/nllb-multi-domain}{NLLB}~\citep{DBLP:journals/corr/abs-2207-04672,DBLP:conf/emnlp/HeffernanCS22}     
        & STS
        & s2s  
        & en
        & 36000 
        & 26504 \\

        \href{https://huggingface.co/datasets/sentence-transformers/embedding-training-data}{Quora}~\citep{quora-question-pairs}   
        & STS
        & s2s  
        & en
        & 92674 
        & 89558 \\

        \href{https://huggingface.co/datasets/multi-train/WikiAnswers_1107}{WikiAnswers}~\citep{DBLP:conf/kdd/FaderZE14}   
        & STS
        & s2s  
        & en
        & 50000 
        & 47686 \\

        \href{https://huggingface.co/datasets/JeremiahZ/simcse_sup_nli}{SimCSE NLI}~\citep{DBLP:conf/emnlp/GaoYC21}   
        & STS
        & s2s  
        & en
        & 252397 
        & 217099 \\

        \href{https://huggingface.co/datasets/stanfordnlp/snli}{SNLI}~\citep{DBLP:conf/emnlp/BowmanAPM15}   
        & STS
        & s2s  
        & en
        & 24686 
        & 16480 \\

        \href{https://huggingface.co/datasets/mteb/raw_arxiv}{arXiv}   
        & Classfication
        & s2s, p2s  
        & en
        & 15000 
        & 14529 \\

        \href{https://huggingface.co/datasets/mteb/raw_biorxiv}{Biorxiv}   
        & Classfication
        & s2s, p2s  
        & en
        & 6862 
        & 6787 \\
        
        \href{https://huggingface.co/datasets/mteb/raw_medrxiv}{Medrxiv}   
        & Classfication
        & s2s, p2s  
        & en
        & 2012 
        & 1999 \\

        \href{https://huggingface.co/datasets/mteb/amazon_polarity}{AmazonPolarity}~\citep{DBLP:conf/recsys/McAuleyL13}
        & Classfication
        & s2s 
        & en
        & 10000 
        & 9007 \\

        \href{https://huggingface.co/datasets/mteb/imdb}{IMDB}~\citep{DBLP:conf/acl/MaasDPHNP11}
        & Classfication
        & s2s 
        & en
        & 10000 
        & 8575 \\

        \href{https://huggingface.co/datasets/mteb/banking77}{banking77}~\citep{DBLP:journals/corr/abs-2003-04807}
        & Classfication
        & s2s 
        & en
        & 10000 
        & 9937 \\

        \href{https://huggingface.co/datasets/mteb/emotion}{EmotionClassification}~\citep{DBLP:conf/emnlp/SaraviaLHWC18}
        & Classfication
        & s2s 
        & en
        & 10000 
        & 10000 \\

        \href{https://huggingface.co/datasets/mteb/tweet_sentiment_extraction}{TweetSentimentExtraction}
        & Classfication
        & s2s 
        & en
        & 10000 
        & 10000 \\

        \href{https://huggingface.co/datasets/mteb/toxic_conversations_50k}{ToxicConversations}
        & Classfication
        & s2s 
        & en
        & 7916 
        & 7800 \\

        \hline

        \href{https://huggingface.co/datasets/C-MTEB/AFQMC}{AFQMC} 
        & STS
        & s2s  
        & zh-cn
        & 4041 
        & 3876 \\

        \href{https://huggingface.co/datasets/shibing624/AdvertiseGen}{AdvertiseGen}~\citep{DBLP:conf/emnlp/ShaoHWXZ19}   
        & Retrieval
        & s2p  
        & zh-cn
        & 20000 
        & 17526 \\

        \href{https://www.luge.ai/#/luge/dataDetail?id=44}{CHEF}~\citep{DBLP:conf/naacl/HuGWLWY22}   
        & Retrieval
        & s2p  
        & zh-cn
        & 4952 
        & 4824 \\

        \href{https://huggingface.co/datasets/michaelwzhu/ChatMed_Consult_Dataset}{ChatMed-Dataset}~\citep{ChatMed}   
        & Retrieval
        & s2p  
        & zh-cn
        & 20000 
        & 18608 \\

        \href{https://huggingface.co/datasets/erhwenkuo/squad-cmrc2018-zhtw}{CMRC 2018}~\citep{DBLP:conf/emnlp/CuiLCXCMWH19}   
        & Retrieval
        & s2p  
        & zh-cn
        & 10000 
        & 9753 \\

        \href{https://huggingface.co/datasets/voidful/DRCD}{DRCD}~\citep{DBLP:journals/corr/abs-1806-00920}   
        & Retrieval
        & s2p  
        & zh-cn
        & 5000 
        & 4714 \\

        \href{https://huggingface.co/datasets/hugcyp/LCSTS}{LCSTS}~\citep{DBLP:conf/emnlp/HuCZ15}   
        & Retrieval
        & s2p  
        & zh-cn
        & 20000 
        & 19535 \\

        \href{https://huggingface.co/datasets/paralym/lima-chinese}{LIMA}~\citep{DBLP:conf/nips/ZhouLX0SMMEYYZG23}   
        & Retrieval
        & s2p  
        & zh-cn
        & 2058 
        & 1991 \\

        \href{https://github.com/Alibaba-NLP/Multi-CPR}{Multi-CPR}~\citep{DBLP:conf/sigir/LongGZXXGXJXY22}   
        & Retrieval
        & s2p  
        & zh-cn
        & 287881 
        & 234587 \\

        \href{https://huggingface.co/datasets/C-MTEB/PAWSX}{PAWS-X (zh)}~\citep{DBLP:conf/emnlp/YangZTB19}   
        & Retrieval
        & s2p  
        & zh-cn
        & 1542 
        & 1542 \\

        \href{https://github.com/sufengniu/RefGPT/blob/main/README_EN.md}{RefGPT}~\citep{DBLP:conf/emnlp/YangYFYWWZ23}   
        & Retrieval
        & s2p  
        & zh-cn
        & 50000 
        & 49896 \\

        \href{https://huggingface.co/datasets/THUIR/T2Ranking}{T2Ranking}~\citep{DBLP:conf/sigir/XieDWLYG0LL0M23}   
        & Retrieval
        & s2p  
        & zh-cn
        & 199412 
        & 188606 \\

        \href{https://huggingface.co/datasets/SirlyDreamer/THUCNews}{THUCNews}~\citep{thuctc}  
        & Retrieval
        & s2p  
        & zh-cn
        & 20000 
        & 19288 \\

        \href{https://www.luge.ai/#/luge/dataDetail?id=62}{UMETRIP-QA}   
        & Retrieval
        & s2p  
        & zh-cn
        & 2647 
        & 2537 \\

        \href{https://github.com/thunlp/WebCPM}{WebCPM}~\citep{DBLP:conf/acl/QinCJYLZLHDWXQL23}   
        & Retrieval
        & s2p  
        & zh-cn
        & 1605 
        & 1602 \\

        \href{https://www.datafountain.cn/competitions/424/datasets}{cCOVID-News}   
        & Retrieval
        & s2p  
        & zh-cn
        & 5000 
        & 4727 \\

        \href{https://huggingface.co/datasets/wangrongsheng/cMedQA-V2.0}{cMedQA-V2.0}~\citep{DBLP:journals/access/ZhangZWGL18}   
        & Retrieval
        & s2p  
        & zh-cn
        & 223851 
        & 88109 \\

        \href{https://huggingface.co/datasets/neuclir/csl}{CSL}~\citep{DBLP:conf/coling/LiZ0S0MZ22}   
        & Retrieval
        & s2p  
        & zh-cn
        & 20000 
        & 19945 \\

        \href{https://huggingface.co/datasets/sentence-transformers/dureader}{DuReader}~\citep{DBLP:conf/acl/HeLLLZXLWWSLWW18}   
        & Retrieval
        & s2p  
        & zh-cn
        & 80416 
        & 79229 \\

        \href{https://huggingface.co/datasets/luozhouyang/dureader}{DuReader\textsubscript{checklist}}~\citep{DBLP:conf/acl/TangL0H0020}   
        & Retrieval
        & s2p  
        & zh-cn
        & 99992 
        & 97764 \\

        \href{https://huggingface.co/datasets/sentence-transformers/law-gpt}{law-gpt}~\citep{LAWGPT-zh}   
        & Retrieval
        & s2p  
        & zh-cn
        & 500 
        & 500 \\

        \href{https://www.heywhale.com/mw/dataset/5e953ca8e7ec38002d02fca7/content}{lawzhidao}~\citep{falv5983}   
        & Retrieval
        & s2p  
        & zh-cn
        & 8000 
        & 6784 \\

        \href{https://huggingface.co/datasets/unicamp-dl/mmarco}{mMARCO (zh)}~\citep{DBLP:journals/corr/abs-2108-13897}  
        & Retrieval
        & s2p  
        & zh-cn
        & 400000 
        & 379870 \\

        \href{https://huggingface.co/datasets/infgrad/retrieval_data_llm}{retrieval\_data\_llm}   
        & Retrieval
        & s2p  
        & zh-cn
        & 32768 
        & 32551 \\

        \href{https://huggingface.co/datasets/suolyer/webqa}{webqa}   
        & Retrieval
        & s2p  
        & zh-cn
        & 5000 
        & 4988 \\

        \href{https://github.com/china-ai-law-challenge/CAIL2019/tree/master/scm}{CAIL2019-SCM}~\citep{DBLP:journals/corr/abs-1911-08962}   
        & STS
        & s2s  
        & zh-cn
        & 5102 
        & 648 \\

        \href{https://www.luge.ai/#/luge/dataDetail?id=39}{CINLID}   
        & STS
        & s2s  
        & zh-cn
        & 5000 
        & 2883 \\

        \href{https://github.com/IAdmireu/ChineseSTS}{ChineseSTS}~\citep{ChineseSTS}   
        & STS
        & s2s  
        & zh-cn
        & 2500 
        & 2497 \\

        \href{https://huggingface.co/datasets/fenffef/cmnli}{CMNLI}~\citep{DBLP:conf/coling/XuHZLCLXSYYTDLS20}   
        & STS
        & s2s  
        & zh-cn
        & 125356 
        & 119029 \\

        \href{https://huggingface.co/datasets/shibing624/nli_zh}{nli\_zh}~\citep{DBLP:conf/emnlp/ChenCLYLT18,DBLP:conf/coling/LiuCDZCLT18,DBLP:conf/emnlp/YangZTB19}   
        & STS
        & s2s  
        & zh-cn
        & 218887 
        & 185787 \\

        \href{https://huggingface.co/datasets/Fred666/ocnli}{OCNLI}~\citep{DBLP:conf/emnlp/HuRXLKM20}   
        & STS
        & s2s  
        & zh-cn
        & 13464 
        & 11937 \\

        \href{https://github.com/CLUEbenchmark/QBQTC/tree/main}{QBQTC}   
        & STS
        & s2s  
        & zh-cn
        & 51620 
        & 47223 \\

        \href{https://github.com/CLUEbenchmark/SimCLUE}{SimCLUE}   
        & STS
        & s2s  
        & zh-cn
        & 344038 
        & 290699 \\
    
        \href{https://huggingface.co/datasets/xnli}{XNLI (zh)}~\citep{DBLP:conf/emnlp/ConneauRLWBSS18}   
        & STS
        & s2s  
        & zh-cn
        & 80000 
        & 74252 \\

        \href{https://huggingface.co/datasets/neuclir/csl}{CSL}~\citep{DBLP:conf/coling/LiZ0S0MZ22}    
        & Classfication
        & s2s, p2s  
        & zh-cn
        & 15000 
        & 12249 \\

        \href{https://huggingface.co/datasets/SirlyDreamer/THUCNews}{THUCNews}~\citep{thuctc}     
        & Classfication
        & s2s  
        & zh-cn
        & 10000 
        & 9690 \\

        \href{https://huggingface.co/datasets/fenffef/tnews}{TNews}
        & Classfication
        & s2s 
        & zh-cn
        & 10000 
        & 6762 \\

        \href{https://huggingface.co/datasets/C-MTEB/JDReview-classification}{JDReview}
        & Classfication
        & s2s  
        & zh-cn
        & 1232 
        & 1232 \\

        \href{https://huggingface.co/datasets/fenffef/iflytek}{IFlyTek}~\citep{DBLP:journals/corr/abs-2202-10974}
        & Classfication
        & s2s 
        & zh-cn
        & 10000 
        & 8221 \\

        \href{https://huggingface.co/datasets/C-MTEB/OnlineShopping-classification}{OnlineShopping}
        & Classfication
        & s2s 
        & zh-cn
        & 7852 
        & 7600 \\

        \href{https://huggingface.co/datasets/C-MTEB/waimai-classification}{Waimai}
        & Classfication
        & s2s 
        & zh-cn
        & 7384 
        & 7376 \\

        \hline

        \href{https://huggingface.co/datasets/CohereForAI/aya_dataset}{Aya Dataset}~\citep{DBLP:conf/acl/SinghVD0MKSPMOZ24}
        & Retrieval
        & s2p  
        & multilingual
        & 30000 
        & 26292 \\

        \href{https://huggingface.co/datasets/sentence-transformers/miracl}{MIRACL}~\citep{DBLP:journals/tacl/0018TOKAL0RL23}   
        & Retrieval
        & s2p  
        & multilingual
        & 40151 
        & 39946 \\

        \href{https://huggingface.co/datasets/castorini/mr-tydi}{Mr. TyDi}~\citep{DBLP:journals/corr/abs-2108-08787}
        & Retrieval
        & s2p  
        & multilingual
        & 48729 
        & 46997 \\

        \href{https://huggingface.co/datasets/maximedb/paws-x-all}{PAWS-X}~\citep{DBLP:conf/emnlp/YangZTB19}   
        & STS
        & s2s  
        & multilingual
        & 128435 
        & 128398 \\

        \href{https://huggingface.co/datasets/mteb/amazon_reviews_multi}{AmazonReviews}~\citep{DBLP:conf/emnlp/NiLM19}
        & Classfication
        & s2s 
        & multilingual
        & 10000 
        & 7721 \\

        \href{https://huggingface.co/datasets/mteb/amazon_counterfactual}{AmazonCounterfactual}~\citep{DBLP:conf/emnlp/ONeillRKKB21}
        & Classfication
        & s2s 
        & multilingual
        & 10000 
        & 8323 \\

        \href{https://huggingface.co/datasets/mteb/multilingual-sentiment-classification}{MultilingualSentiment}~\citep{mollanorozy2023cross}
        & Classfication
        & s2s 
        & multilingual
        & 10000 
        & 9804 \\

        \href{https://huggingface.co/datasets/mteb/amazon_massive_intent}{Amazon Massive Intent}~\citep{DBLP:conf/acl/FitzGeraldHPMRS23}
        & Classfication
        & s2s 
        & multilingual
        & 10000 
        & 7832 \\

        \href{https://huggingface.co/datasets/mteb/amazon_massive_scenario}{AmazonMassiveScenario}~\citep{DBLP:conf/acl/FitzGeraldHPMRS23}
        & Classfication
        & s2s 
        & multilingual
        & 10000 
        & 7078 \\

        \href{https://huggingface.co/datasets/mteb/mtop_domain}{MTOPDomain}~\citep{DBLP:conf/eacl/LiACGGM21}
        & Classfication
        & s2s 
        & multilingual
        & 10000 
        & 9610 \\

        \href{https://huggingface.co/datasets/mteb/mtop_intent}{MTOPIntent}~\citep{DBLP:conf/eacl/LiACGGM21}
        & Classfication
        & s2s 
        & multilingual
        & 10000 
        & 7952 \\
        
        \bottomrule
    \end{tabular}

    \caption{Fine-tuning data list.}
    \label{tab:Fine-tuning_data_list}
\end{table}

\newpage
\begin{table}[htbp]
  \centering
  \renewcommand{\arraystretch}{1.395}
  \tiny
  \begin{tabular}{lp{9.7cm}}
    \toprule
    \textbf{Task Name }     &  \textbf{Instruction}  \\
    
    \hline
    \multicolumn{2}{c}{\textbf{Classification}} \\
    \hline
    
    AmazonCounterfactualClassification & Instruct: Given an Amazon review, judge whether it is counterfactual. \textbackslash n Query: \{query\} \\
    \hline
    
    AmazonPolarityClassification & Instruct: Classifying Amazon reviews into positive or negative sentiment \textbackslash n Query: \{query\} \\
    \hline
    
    AmazonReviewsClassification & Instruct: Classifying the given Amazon review into its appropriate rating category \textbackslash n Query: \{query\} \\
    \hline
    
    Banking77Classification & Instruct: Given a online banking query, find the corresponding intents \textbackslash n Query: \{query\} \\
    \hline
    
    EmotionClassification & Instruct: Classifying the emotion expressed in the given Twitter message into one of the six emotions: anger, fear, joy, love, sadness, and surprise \textbackslash n Query: \{query\} \\
    \hline
    
    ImdbClassification & Instruct: Classifying the sentiment expressed in the given movie review text from the IMDB dataset \textbackslash n Query: \{query\} \\
    \hline
    
    MassiveIntentClassification & Instruct: Given a user utterance as query, find the user intents \textbackslash n Query: \{query\} \\
    \hline
    
    MassiveScenarioClassification & Instruct: Given a user utterance as query, find the user scenarios \textbackslash n Query: \{query\} \\
    \hline
    
    MTOPDomainClassification & Instruct: Classifying the intent domain of the given utterance in task-oriented conversation \textbackslash n Query: \{query\} \\
    \hline
    
    MTOPIntentClassification & Instruct: Classifying the intent of the given utterance in task-oriented conversation \textbackslash n Query: \{query\} \\
    \hline
    
    ToxicConversationsClassification & Instruct: Classifying the given comments as either toxic or not toxic \textbackslash n Query: \{query\} \\
    \hline
    
    TweetSentimentExtractionClassification & Instruct: Classifying the sentiment of a given tweet as either positive, negative, or neutral \textbackslash n Query: \{query\} \\
    \hline
    
    TNews & Instruct: Categorizing the given news title \textbackslash n Query: \{query\} \\
    \hline
    
    IFlyTek & Instruct: Given an App description text, find the appropriate fine-grained category \textbackslash n Query: \{query\} \\
    \hline
    
    MultilingualSentiment & Instruct: Classifying sentiment of the customer review into positive, neutral, or negative \textbackslash n Query: \{query\} \\
    \hline
    
    JDReview & Instruct: Classifying sentiment of the customer review for iPhoneinto positive or negative \textbackslash n Query: \{query\} \\
    \hline
    
    OnlineShopping & Instruct: Classifying sentiment of the customer reviewinto positive or negative \textbackslash n Query: \{query\} \\
    \hline
    
    Waimai & Instruct: Classify the customer review from a food takeaway platform into positive or negative \textbackslash n Query: \{query\} \\
    \hline
    
    MasakhaNEWSClassification & Instruct: Classifying the category of french news. \textbackslash n Query: \{query\} \\
    \hline
    
    CBD & Instruct: Classifying the sentiment of polish tweet reviews \textbackslash n Query: \{query\} \\
    \hline
    
    PolEmo2.0-IN & Instruct: Classifying the sentiment of in-domain (medicine and hotels) online reviews \textbackslash n Query: \{query\} \\
    \hline
    
    PolEmo2.0-OUT & Instruct: Classifying the sentiment of out-of-domain (products and school) online reviews \textbackslash n Query: \{query\} \\
    \hline
    
    AllegroReviews & Instruct: Classifying the sentiment of reviews from e-commerce marketplace Allegro \textbackslash n Query: \{query\} \\
    \hline
    
    PAC & Instruct: Classifying the sentence into one of the two types: "BEZPIECZNE\_POSTANOWIENIE\_UMOWNE" and "KLAUZULA\_ABUZYWNA" \textbackslash n Query: \{query\} \\
    \hline
    
    GeoreviewClassification & Instruct: Classifying the sentiment of Russian reviews. \textbackslash n Query: \{query\} \\
    \hline
    
    HeadlineClassification & Instruct: Classifying the topic of Russian headlines. \textbackslash n Query: \{query\} \\
    \hline
    
    InappropriatenessClassification & Instruct: Detecting inappropriate messages on sensitive topics \textbackslash n Query: \{query\} \\
    \hline
    
    KinopoiskClassification & Instruct: Classifying the sentiment of Kinopoisk reviews. \textbackslash n Query: \{query\} \\
    \hline
    
    RuReviewsClassification & Instruct: Classifying the sentiment of Russian product reviews. \textbackslash n Query: \{query\} \\
    \hline
    
    RuSciBenchGRNTIClassification & Instruct: Classifying the topic of Russian scientific papers. \textbackslash n Query: \{query\} \\
    \hline
    
    RuSciBenchOECDClassification & Instruct: Classifying the topic of Russian scientific papers. \textbackslash n Query: \{query\} \\
    \hline
    
    CEDRClassification & Instruct: Classification of sentences by emotions. \textbackslash n Query: \{query\} \\
    \hline
    
    SensitiveTopicsClassification & Instruct: Detecting inappropriate messages on sensitive topics. \textbackslash n Query: \{query\} \\

    \hline
    \multicolumn{2}{c}{\textbf{Clustering}} \\
    \hline

    ArxivClusteringP2P & Instruct: Identify the main and secondary category of Arxiv papers based on the titles and abstracts \textbackslash n Query: \{query\} \\
    \hline
    
    ArxivClusteringS2S & Instruct: Identify the main and secondary category of Arxiv papers based on the titles \textbackslash n Query: \{query\} \\
    \hline
    
    BiorxivClusteringP2P & Instruct: Identify the main category of Biorxiv papers based on the titles and abstracts \textbackslash n Query: \{query\} \\
    \hline
    
    BiorxivClusteringS2S & Instruct: Identify the main category of Biorxiv papers based on the titles \textbackslash n Query: \{query\} \\
    \hline
    
    MedrxivClusteringP2P & Instruct: Identify the main category of Medrxiv papers based on the titles and abstracts \textbackslash n Query: \{query\} \\
    \hline
    
    MedrxivClusteringS2S & Instruct: Identify the main category of Medrxiv papers based on the titles \textbackslash n Query: \{query\} \\
    \hline
    
    RedditClustering & Instruct: Identify the topic or theme of Reddit posts based on the titles \textbackslash n Query: \{query\} \\
    \hline
    
    RedditClusteringP2P & Instruct: Identify the topic or theme of Reddit posts based on the titles and posts \textbackslash n Query: \{query\} \\
    \hline
    
    StackExchangeClustering & Instruct: Identify the topic or theme of StackExchange posts based on the titles \textbackslash n Query: \{query\} \\
    \hline
    
    StackExchangeClusteringP2P & Instruct: Identify the topic or theme of StackExchange posts based on the given paragraphs \textbackslash n Query: \{query\} \\
    \hline
    
    TwentyNewsgroupsClustering & Instruct: Identify the topic or theme of the given news articles \textbackslash n Query: \{query\} \\
    \hline
    
    CLSClusteringS2S & Instruct: Identify the main category of scholar papers based on the titles \textbackslash n Query: \{query\} \\
    \hline
    
    CLSClusteringP2P & Instruct: Identify the main category of scholar papers based on the titles and abstracts \textbackslash n Query: \{query\} \\
    \hline
    
    ThuNewsClusteringS2S & Instruct: Identify the topic or theme of the given news articles based on the titles \textbackslash n Query: \{query\} \\
    \hline
    
    ThuNewsClusteringP2P & Instruct: Identify the topic or theme of the given news articles based on the titles and contents \textbackslash n Query: \{query\} \\
    \hline
    
    AlloProfClusteringP2P & Instruct: Identify the main category of Allo Prof document based on the titles and descriptions \textbackslash n Query: \{query\} \\
    \hline
    
    AlloProfClusteringS2S & Instruct: Identify the main category of Allo Prof document based on the titles \textbackslash n Query: \{query\} \\
    \hline
    
    HALClusteringS2S & Instruct: Identify the main category of academic passage based on the titles and contents \textbackslash n Query: \{query\} \\
    \hline
    
    MasakhaNEWSClusteringP2P & Instruct: Identify the topic or theme of the given news articles based on the titles and contents \textbackslash n Query: \{query\} \\
    \hline
    
    MasakhaNEWSClusteringS2S & Instruct: Identify the topic or theme of the given news articles based on the titles \textbackslash n Query: \{query\} \\
    \hline
    
    MLSUMClusteringP2P & Instruct: Identify the topic or theme of the given articles based on the titles and contents \textbackslash n Query: \{query\} \\
    \hline
    
    MLSUMClusteringS2S & Instruct: Identify the topic or theme of the given articles based on the titles \textbackslash n Query: \{query\} \\
    \hline
    
    EightTagsClustering & Instruct: Identify of headlines from social media posts in Polish  into 8 categories: film, history, food, medicine, motorization, work, sport and technology \textbackslash n Query: \{query\} \\
    \hline
    
    GeoreviewClusteringP2P & Instruct: Identify the topic or theme of the Russian reviews. \textbackslash n Query: \{query\} \\
    \hline
    
    RuSciBenchGRNTIClusteringP2P & Instruct: Identify the topic or theme of the Russian articles. \textbackslash n Query: \{query\} \\
    \hline
    
    RuSciBenchOECDClusteringP2P & Instruct: Identify the topic or theme of the Russian articles. \textbackslash n Query: \{query\} \\

    \bottomrule
  \end{tabular}
  \caption{Detailed task instruction list for MTEB evaluation.}
  \label{tab:task_instruction_detailed_list}
\end{table}

\end{document}